\documentclass[10pt,twocolumn,letterpaper]{article}
\usepackage{nopageno}
\pdfoutput=1

\usepackage{color}
\usepackage[dvipsnames]{xcolor}
\usepackage{tikz}
\usetikzlibrary{shapes.callouts}

\usepackage{cvpr}

\usepackage{times}
\usepackage{relsize}
\usepackage[T1]{fontenc} 
\usepackage{inconsolata}
\usepackage[latin1]{inputenc}
\usepackage[english]{babel}

\usepackage[normalem]{ulem}
\usepackage{epsfig}
\usepackage{graphicx}
\usepackage{wrapfig}
\usepackage[belowskip=0pt,aboveskip=3pt,font=small]{caption}
\usepackage[belowskip=0pt,aboveskip=0pt,font=small]{subcaption}
\usepackage{amsmath, amsthm, amssymb}
\usepackage{textcomp}
\usepackage{stmaryrd}
\usepackage{upgreek}
\usepackage{bm}
\usepackage{cases}
\usepackage{mathtools}
\usepackage{arydshln}
\usepackage{multirow}
\usepackage{makecell}
\usepackage{graphbox}

\usepackage{cite}
\usepackage[pagebackref=true,breaklinks=true,letterpaper=true,colorlinks,bookmarks=false,citecolor=green]{hyperref}

\usepackage{bibspacing}
\usepackage{algorithm}
\usepackage{algpseudocode}

\usepackage{multirow}
\usepackage{rotating}
\usepackage{booktabs}
\usepackage{tabularx}

\usepackage{enumitem}
\usepackage[olditem,oldenum]{paralist}

\usepackage{alltt}
\usepackage{listings}

\belowdisplayskip \abovedisplayskip

\newlength{\sectionReduceTop}
\newlength{\sectionReduceBot}
\newlength{\subsectionReduceTop}
\newlength{\subsectionReduceBot}
\newlength{\abstractReduceTop}
\newlength{\abstractReduceBot}
\newlength{\captionReduceTop}
\newlength{\captionReduceBot}
\newlength{\subsubsectionReduceTop}
\newlength{\subsubsectionReduceBot}

\newlength{\eqnReduceTop}
\newlength{\eqnReduceBot}

\newlength{\horSkip}
\newlength{\verSkip}

\newlength{\figureHeight}
\usepackage{mysymbols}

\usepackage{url}
\usepackage{xspace}
\usepackage{comment}

\usepackage{afterpage}
\usepackage{pdfpages}
\usepackage{framed}
\usepackage{fancybox}
\usepackage{cuted}
\usepackage{caption}

\usepackage{enumitem}

\setdefaultleftmargin{.2cm}{.2cm}{}{}{}{}
\setlist{leftmargin=.2cm}

\newcommand{\Dialog}{Dialog\xspace}

\newcommand{\xhdr}[1]{\vspace{1pt}\noindent\textbf{#1}}

\ifcvprfinal\fi
\cvprfinalcopy 

\def\cvprPaperID{3469} 

\usepackage{flushend}

\begin{document}

\title{Audio Visual Scene-Aware \Dialog}

\author{
    Huda Alamri$^1$, Vincent Cartillier$^1$, Abhishek Das$^1$, Jue Wang$^2$ ,
    Anoop Cherian$^2$, Irfan Essa$^1$,\\
    Dhruv Batra$^1$, Tim K. Marks$^2$, Chiori Hori$^2$, 
    Peter Anderson$^1$, Stefan Lee$^1$, Devi Parikh$^1$ \\
    $^1$Georgia Institute of Technology \quad
    $^2$Mitsubishi Electric Research Laboratories (MERL) \\
    {\tt\small $^1$\{halamri, vcartillier3, abhshkdz, irfan, dbatra, peter.anderson, steflee, parikh\}@gatech.edu} \\
    {\tt\small $^2$\{juewangj, cherian, tmarks, chori\}@merl.com} \\
    \tt\normalsize
    \href{https://video-dialog.com}{video-dialog.com}
}

\maketitle

\vspace{\abstractReduceTop}
\begin{abstract}
\vspace{\abstractReduceBot}
\vspace{-2pt}

We introduce the task of scene-aware dialog. Our goal is to generate a complete and natural response to a question about a scene, given video and audio of the scene and the history of previous turns in the dialog. To answer successfully, agents must ground concepts from the question in the video while leveraging contextual cues from the dialog history. To benchmark this task, we introduce the Audio Visual Scene-Aware Dialog (AVSD) Dataset. For each of more than 11,000 videos of human actions from the Charades dataset, our dataset contains a dialog about the video, plus a final summary of the video by one of the dialog participants. We train several baseline systems for this task and evaluate the performance of the trained models using both qualitative and quantitative metrics. Our results indicate that models must utilize all the available inputs (video, audio, question, and dialog history) to perform best on this dataset. 
\vspace{-3pt}
\end{abstract}

\vspace{\sectionReduceTop}
\section{Introduction}
\label{sec:intro}

\begin{figure}[t]
    \centering
    \includegraphics[width=0.95\columnwidth]{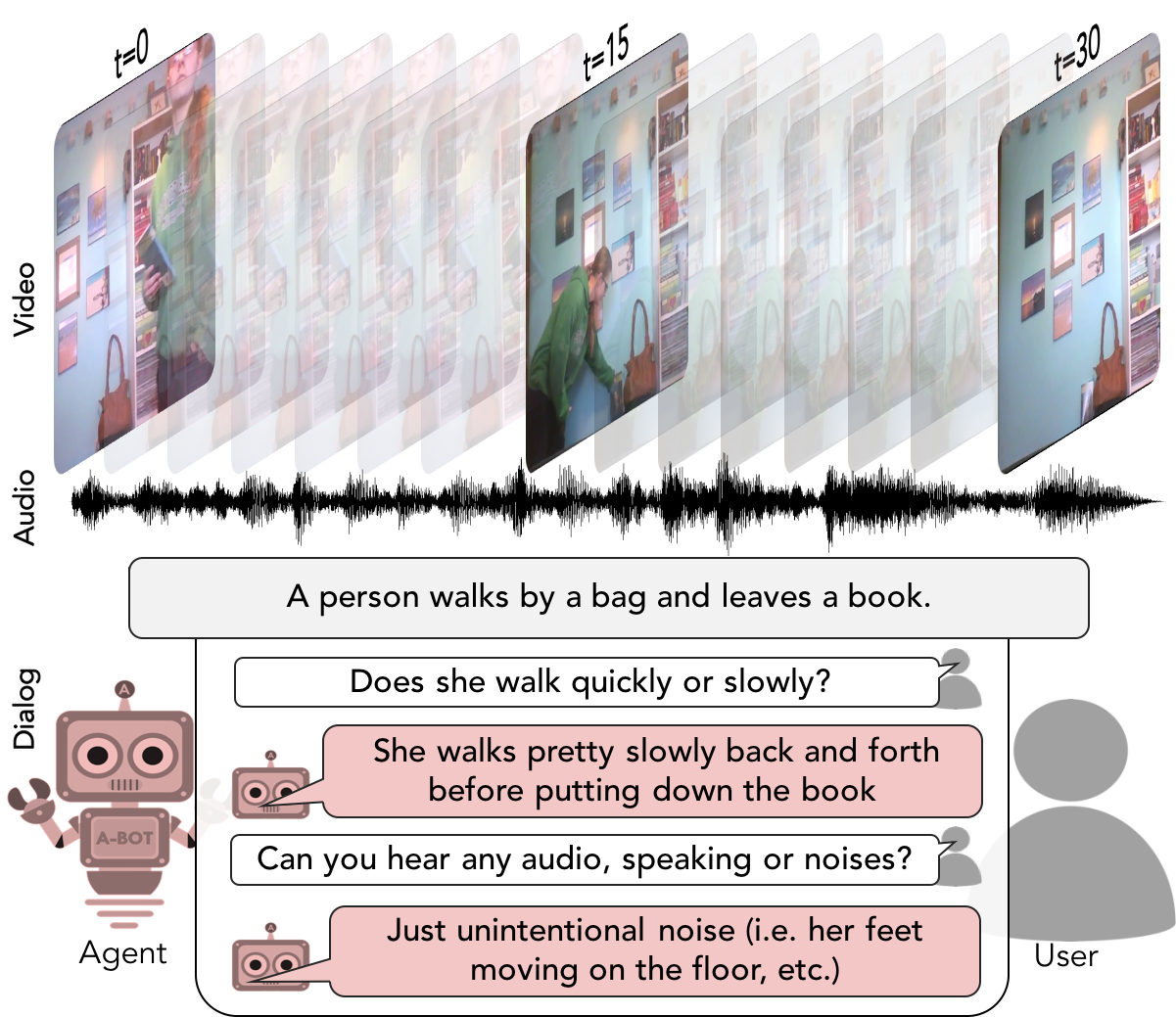}
    \caption{In Audio Visual Scene-Aware Dialog, an agent's task is to answer natural language questions about a short video. The agent grounds its responses on the dynamic scene, the audio, and the history (previous rounds) of the dialog,
    dialog history, which begins with a short script of the scene.}
    \vspace{-10pt}
    \label{fig:teaser}
\end{figure}

Developing conversational agents has been a longstanding goal of artificial intelligence (AI). For many human-computer interactions, natural language presents an ideal interface, as it is fully expressive and requires no user training. One emerging area is the development of {\em visually aware} dialog systems. Das \etal~\cite{das2017visual} introduced the problem of {\em visual dialog}, in which a model is trained to carry out a conversation in natural language, answering questions about an image. For a given question, the system has to ground its response in the input image as well as the previous utterances. 
However, conversing about a static image is inherently limiting. Many potential applications for conversational agents, such as virtual personal assistants and assistive technologies for the visually impaired, would benefit greatly from understanding the scene in which the agent is operating. This context often cannot be captured in a single image, as there is important information in the temporal dynamics of the scene as well as in the audio. 

Our goal is to move towards conversational agents that are not only visually intelligent but also aware of temporal dynamics.
For example, a security guard (G) might have the following exchange with an AI agent: 
``G: Has there been anyone carrying a red handbag in the last week in Sector 5? AI: Yes, a woman in a blue suit. G: Do any of the exit cameras show her leaving with it? AI: No. G: Did anyone else pick it up?''.

Answering such questions requires a holistic understanding of the visual and audio information in the scene, including temporal dynamics. Since human communication is rarely single-shot, an understanding of sequential dialog (e.g., what {\em her} and {\em it} refer to) is also required.

We introduce the task of scene-aware dialog, as well as the Audio Visual Scene-aware Dialog (AVSD) Dataset to provide a means for training and testing scene-aware dialog systems. In the general task of scene-aware dialog, the goal of the system is to carry on a conversation with a human about a temporally varying scene. 
In the AVSD Dataset, we are addressing a particular type of scene-aware dialog. Each dialog in the dataset is a sequence of question/answer (QA) pairs about a short video; each video features a person performing everyday activities in a home environment. 

We defined a specific task for the scene-aware dialog system to learn: Given an input video, the history of a dialog about the video (consisting of a short script plus the first $t\!-\!1$ QA pairs), and a follow-up question (the $t$th question in the dialog), the system's goal is to generate a correct response to the follow-up question.
We aim to use the dataset to explore the compositionality of dynamic scenes and to train an end-to-end model to leverage information from the video frames, audio signals, and dialog history. The system should engage in this conversation by providing complete and natural responses to enable real-world applicability. 
The development of such scene-aware conversational agents represents an important frontier in artificial intelligence. In addition, it holds promise for numerous practical applications, such as video retrieval from users' free-form queries, and helping visually impaired people understand visual content.
Our contributions include the following:

\begin{enumerate}[topsep=0pt, partopsep=0pt, itemsep=2pt, parsep=0pt, leftmargin=14pt]
    \item We introduce the task of scene-aware dialog, which is a multimodal semantic comprehension task.
    \item We introduce a new benchmark for the scene-aware dialog task, the AVSD Dataset, consisting of more than 11,000 conversations that discuss the content (including actions, interactions, sound, and temporal dynamics) of videos of human-centered activities. 
    \item We analyze the performance of several baseline systems on this new benchmark dataset.
\end{enumerate}

\section{Related Work}
\label{sec:related}

\textbf{Video Datasets:}
In the domain of dynamic scene understanding, a large body of literature focuses on action classification.
Some important benchmarks for video-action recognition and detection are: HMDB51~\cite{kuehne2011hmdb} 
Sports-1M~\cite{karpathy2014large} and UCF-101~\cite{soomro2012ucf101}. The ActivityNet~\cite{caba2015activitynet} and Kinetics~\cite{kay2017kinetics} datasets target a broader range of human-centered action categories. Sigurdsson \etal \cite{sigurdsson2016hollywood} presented the Charades dataset. Charades is a crowd-sourced video dataset that was built by asking Amazon Mechanical Turk (AMT) workers to write some scene scripts of daily activities, then asking another group of AMT workers to record themselves ``acting out'' the scripts in a ``Hollywood style.'' In our work, Charades videos were used to collect conversations about the activities in the videos. Video-based captioning is another exciting research area, and there are several datasets introduced to benchmark and evaluate this task  \cite{krishna2017dense, rohrbach2014coherent}.

\textbf{Video-based Question Answering:} Inspired by the success of image-based question answering \cite{antol2015vqa, yu2015visual, gao2015you, zhu2016visual7w}, recent work has addressed the task of video-based question answering \cite{maharaj2017dataset, jang2017tgif, Tapaswi_2016}. MovieQA \cite{lei2018tvqa} and TVQA \cite{Tapaswi_2016} evaluate the video-based QA task by training end-to-end models to answer multiple-choice questions by leveraging cues from video frames and associated textual information, such as scripts and subtitles. While this one-shot question answering setting is more typical in existing work, we find this structure to be unnatural. Our focus in AVSD is on settings involving multiple rounds of questions that require natural free-form answers.

\textbf{Visual Dialog:} Our work is directly related to the image-based dialog task (VisDial) introduced by Das \etal \cite{das2017visual}. Given an input image, a dialog history, and a question, the agent is required to answer the question while grounding the answer on the input image and the dialog history. In \cite{das2017visual}, several network architectures are introduced to encode the different input modalities: late fusion, hierarchical recurrent encoder, and memory network. In this work, we extend the work from \cite{das2017visual} to include additional complex modalities: video frames and audio signals.
\vspace{-5pt}
\begin{table}[b]
\centering 
\resizebox{1\columnwidth}{!}{
\footnotesize
\setlength{\tabcolsep}{0.3em}
\begin{tabular}{l c c c c} 
\toprule
Dataset & \# Video Clips & \# QA Pairs & Video Source & Answers \\  
\midrule %
TVQA\cite{lei2018tvqa} &  21,793  &  152,545 &  TV shows  & Multi-Choice \\ 
MovieQA\cite{Tapaswi_2016} &  408  & 14,944  &  Movies &  Multi-Choice \\
TGIF-QA\cite{jang2017tgif} &  56,720 &  103,919 &  Social media & Multi-Choice\\
VisDial\cite{das2017visual} & 120K (images) & 1.2 M & N/A & Free-Form\\
AVSD (Ours) & 11,816  & 118,160  & Crowdsourced & Free-Form\\
\bottomrule %
\end{tabular}}
\caption{Comparison with existing video question answering and visual dialog datasets.}
\vspace{-10pt}
\label{table:DatasetsComparison} 
\end{table}

\section{Audio Visual Scene-Aware Dialog Dataset}
\label{sec:dataset}

A primary goal of our paper is to create a benchmark for the task of scene-aware dialog. There are several characteristics that we desire for such a dataset:
1) The dialogs should focus on the dynamic aspects of the video (i.e., actions and interactions);
2) The answers should be complete explanatory responses rather than brief one- or two-word answers (e.g., not simply yes or no);
3) The dialogs should discuss the temporal order of events in the video. 

\begin{figure}
    \centering
    \footnotesize
    \setlength{\tabcolsep}{0.3em}
    \renewcommand{\arraystretch}{2}
    \begin{tabularx}{\linewidth}{lX}
        \includegraphics[align=t,vshift=0.1cm,width=0.32\columnwidth, height=2cm]{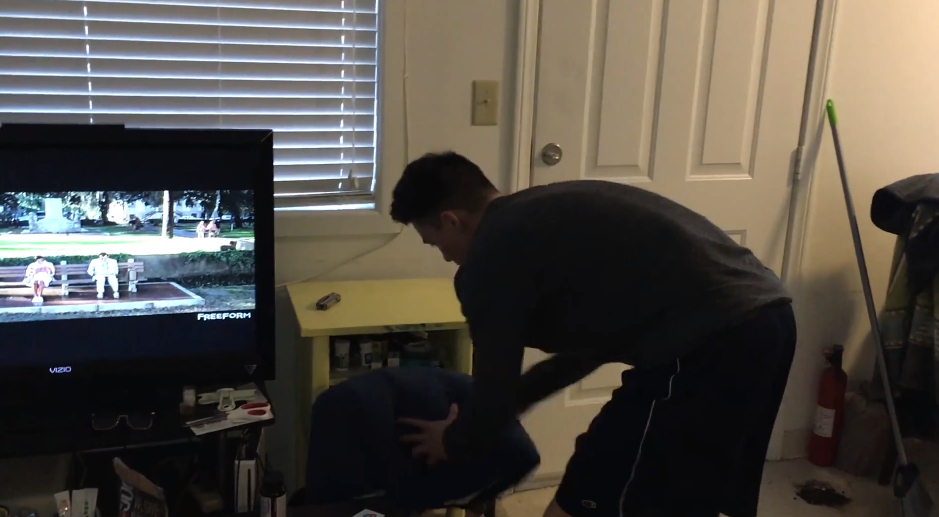} &  A person is throwing a pillow into the wardrobe. Then, taking the dishes off the table, the person begins tidying up the room.\\
    \end{tabularx}
    \begin{tabularx}{\linewidth}{lX}
        \includegraphics[align=t,vshift=0.1cm,width=0.32\columnwidth, height=1.8cm]{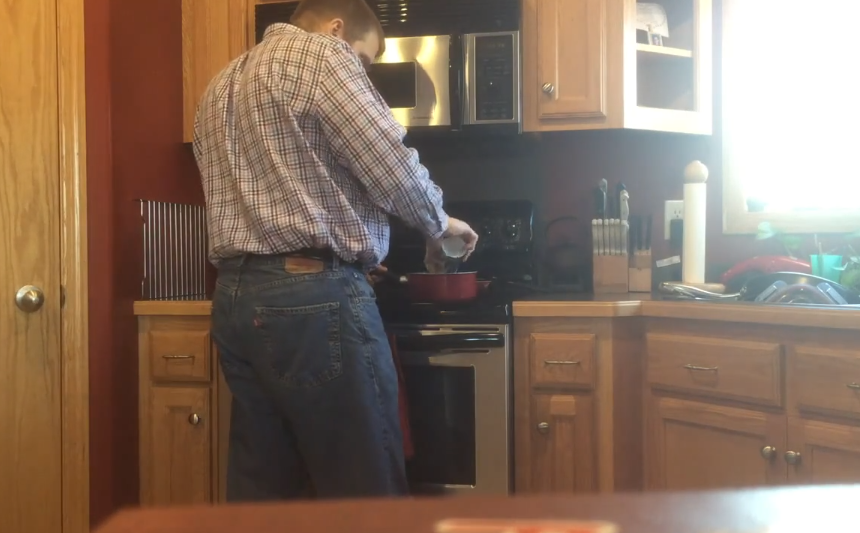} & A person is pouring some liquid into a pot as they cook at a stove. They open a cabinet and take out a picture, and set it next to the stove while they continue to cook and gaze at the photo.\\
    \end{tabularx}
    \begin{tabularx}{\linewidth}{lX}
        \includegraphics[align=t,vshift=0.1cm,width=0.32\columnwidth, height=2.3cm]{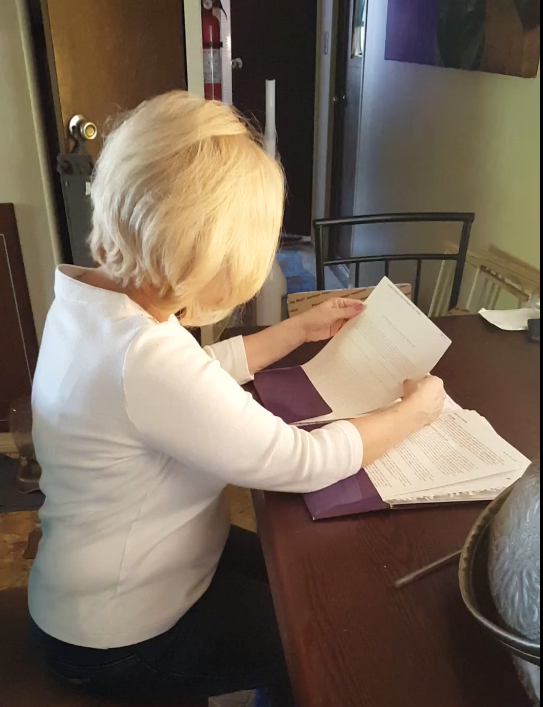} & The person leaves their homework at the table as they get up and rub their stomach to indicate hunger. The person walks towards the pantry and grabs the doorknob. After twisting the knob and opening the door, the person is disappointed to find canned food and stacks of phone books.\\
    \end{tabularx}
    \caption{Examples of videos and scripts from the Charades~\cite{sigurdsson2016hollywood} dataset. Each video's temporally ordered sequence of small events is a good fit for our goal to train a video-based dialog system.}
    \vspace{-10pt}
    \label{fig:Charades} 
\end{figure}

\xhdr{Video Content.}
An essential element to collecting video-grounded dialogs is of course the videos themselves. We choose to collect dialogs grounded in the Charades~\cite{sigurdsson2016hollywood} human-activity dataset. The Charades dataset consists of 11,816 videos of everyday indoor human activities with an average length of 30 seconds. Each video includes at least two actions. Examples of frames and scripts for Charades videos are shown in Figure~\ref{fig:Charades}. We choose the Charades dataset for two main reasons. First, the videos in this dataset are crowd-sourced on Amazon Mechanical Turk (AMT), so the settings are natural and diverse. Second, each video consists of a sequence of small events that provide AMT Workers (Turkers) with rich content to discuss.


\vspace{\subsectionReduceTop}
\subsection{Data Collection}
\label{sec:data_collection}
We adapt the real-time chat interface from \cite{das2017visual} to pair two AMT workers to have an English-language conversation about a video from the Charades Dataset (Figure~\ref{fig:Charades}). One person, the ``Answerer,'' is presented with the video clip and the script, and their role is to provide detailed answers to questions about the scene. The other person, the ``Questioner,'' does not have access to the video or the script, and can only see three frames (one each from the beginning, middle, and end) of the video. The Questioner's goal is to ask questions to obtain a good understanding of what happens in the video scene. We considered several design choices for the chat interface and instructions, in order to encourage natural conversations about events in the videos.

\xhdr{Investigating Events in Video.} 
To help distinguish this task from previous image and video captioning tasks, 
our instructions direct the Questioner to ``investigate  what is happening'' rather than simply asking the two Turkers to ``chat about the video.'' We find that when asked to ``chat about the video,'' Questioners tend to ask a lot of questions about the setting and the appearance of the people in the video. In contrast, the direction ``investigate  what is happening'' leads Questioners to inquire more about the actions of the people in the video.

\begin{figure}
\begin{center}
\includegraphics[width=1\linewidth, clip]{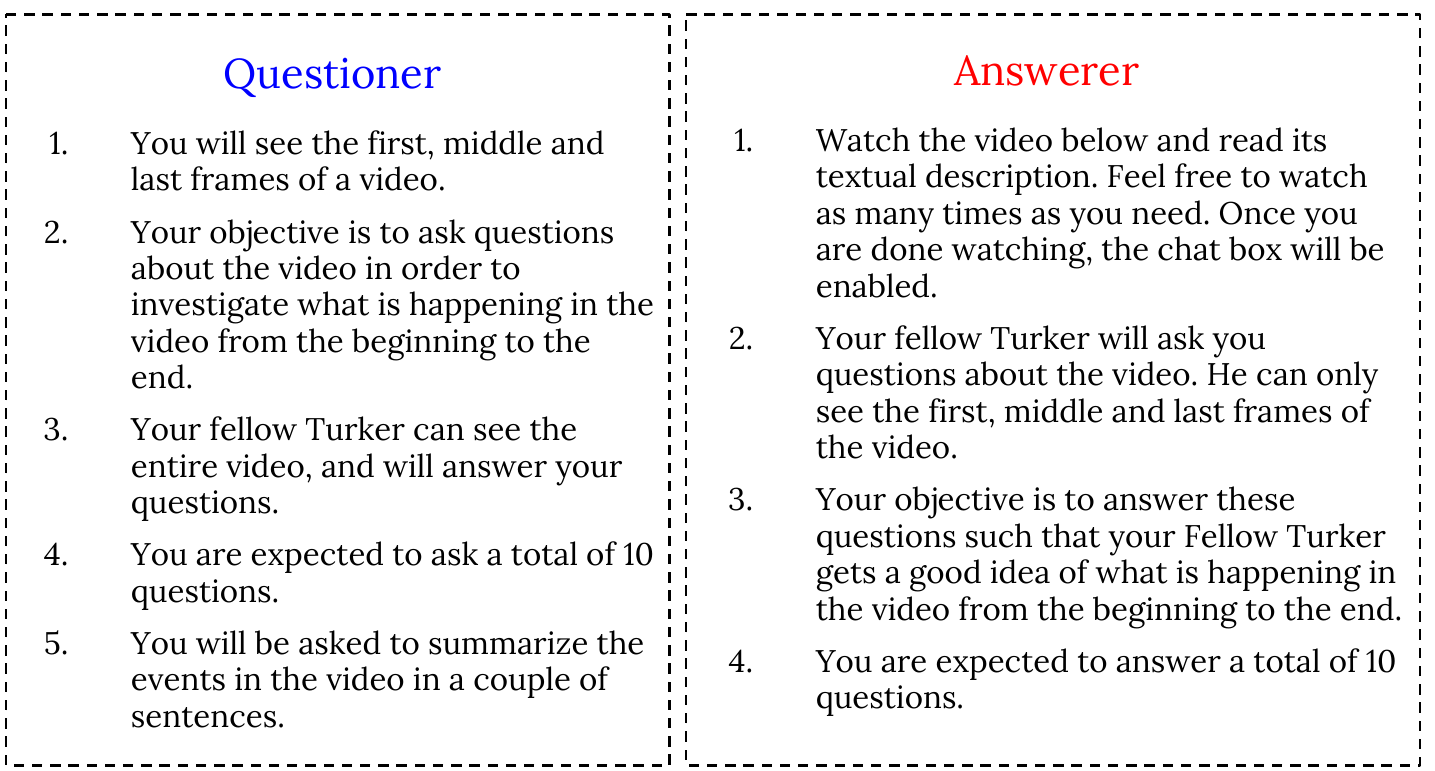}
\end{center}
\caption{Instructions provided to AMT workers explaining the roles of ``Questioner'' and ``Answerer.''}
\vspace{-10pt}
\label{fig:AMTinstructions}
\end{figure}

\xhdr{Seeding the Conversation.} 
There are two reasons that our protocol provides the Questioners with three frames before the conversation starts: First, since the images provide the overall layout of the scene, they ensure that the conversations are centered around the actions and  events that take place in the video rather than about the scene layout or the appearance of people and objects. Second, we found that providing multiple frames instead of a single frame encouraged users to ask about the sequence of events. 
Providing the Questioners with these three images achieves both criteria without explicitly dictating Questioners' behavior; this is important because we want the conversations to be as natural as possible. 

\vspace{-5pt}

\xhdr{Downstream Task: Video Summarization.} 
Once the conversation (sequence of 10 QA pairs) between the Questioner and Answerer is complete, the Questioner's final task is to summarize what they think happened in the video. Knowing that this will be their final task motivates the Questioner to ask good questions that will lead to informative answers about the events in the video. In addition, this final downstream task is used to evaluate the quality of the dialog and how informative it was about the video.
Figure \ref{fig:AMTinstructions} shows the list of instructions provided to AMT workers.

\xhdr{Worker Qualifications.} To ensure high-quality and fluent dialogs, we restrict our tasks on AMT to Turkers with $\geq 95\%$ task acceptance rates, located in North America, who have completed at least 500 tasks already. We further restrict any one Turker from completing more than 200 tasks in order to maintain diversity. In total, 1553 unique workers contributed to the dataset collection effort.

Table \ref{table:DatasetsComparison} puts the Audio Visual Scene-aware Dialog (AVSD) Dataset in context with several other video question answering benchmarks. While AVSD has fewer unique video clips compared to TVQA and MovieQA, which are curated from television and film, our videos are more naturalistic. Moreover, AVSD contains a similar number of questions and answers, but as a part of multi-round dialogs.

\begin{figure}
\begin{center}
\includegraphics[width=1\linewidth]{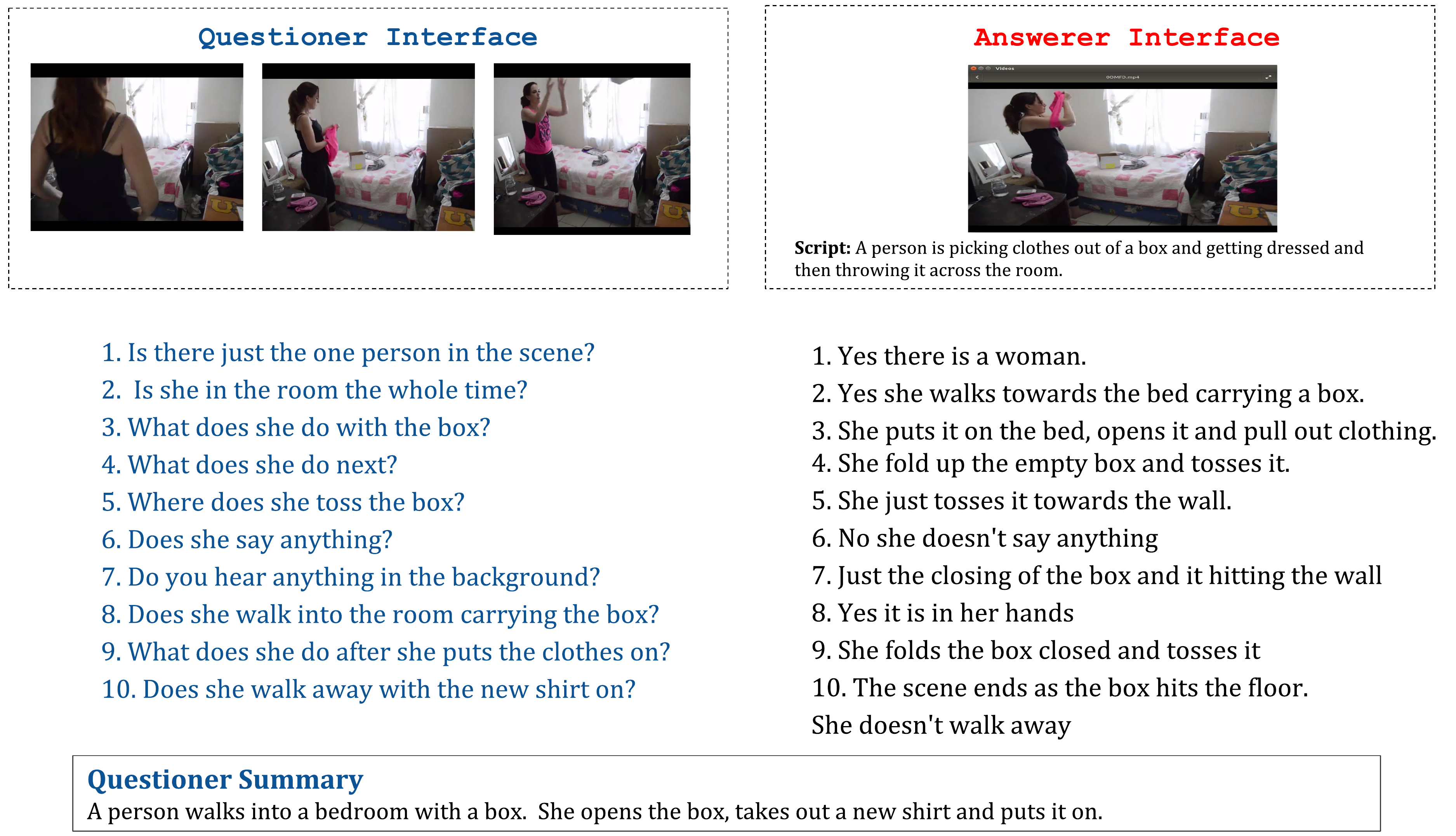}
\end{center}
\vspace{-5pt}
\caption{Example conversation between two AMT workers. The Questioner is presented with 3 static images from the video and asks a question. The Answerer, who has already watched the video and read the script, responds. After 10 rounds of QA, the Questioners provides a written summary of what they think happened in the video based on the conversation.}
\vspace{-10pt}
\label{fig:conversation}
\end{figure}


\vspace{\subsectionReduceTop}
\subsection{AVSD Dataset Analysis}
\vspace{\subsectionReduceBot}
\label{sec:data_analysis}

\begin{figure*}[t]
    \centering
    \begin{subfigure}[t]{0.35\textwidth}
    \includegraphics[width=\linewidth]{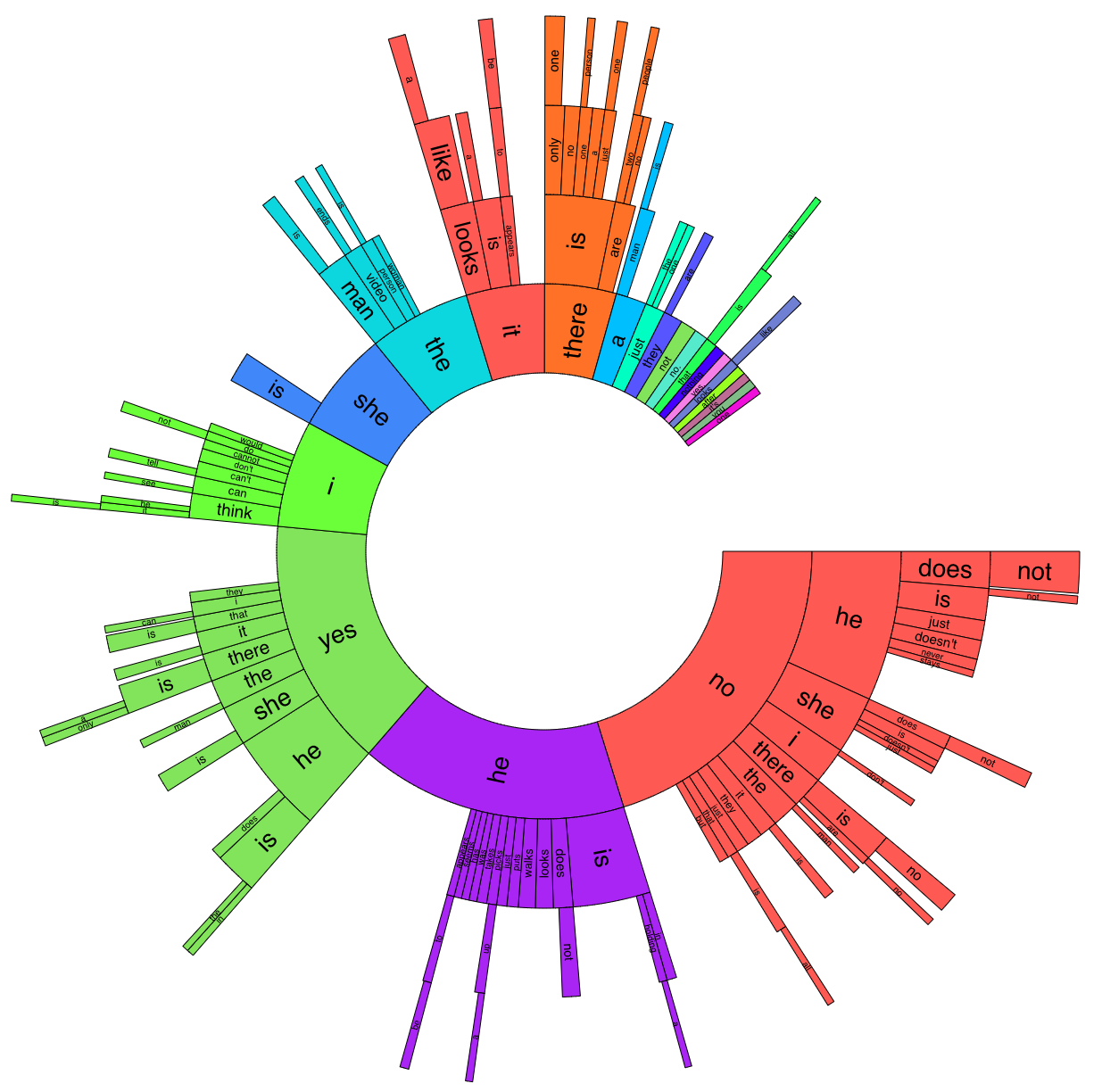}
    \caption{AVSD Answers}
    \label{fig:avsdAns}
	\end{subfigure}
	\begin{subfigure}[t]{0.35\textwidth}
    \includegraphics[width=\linewidth]{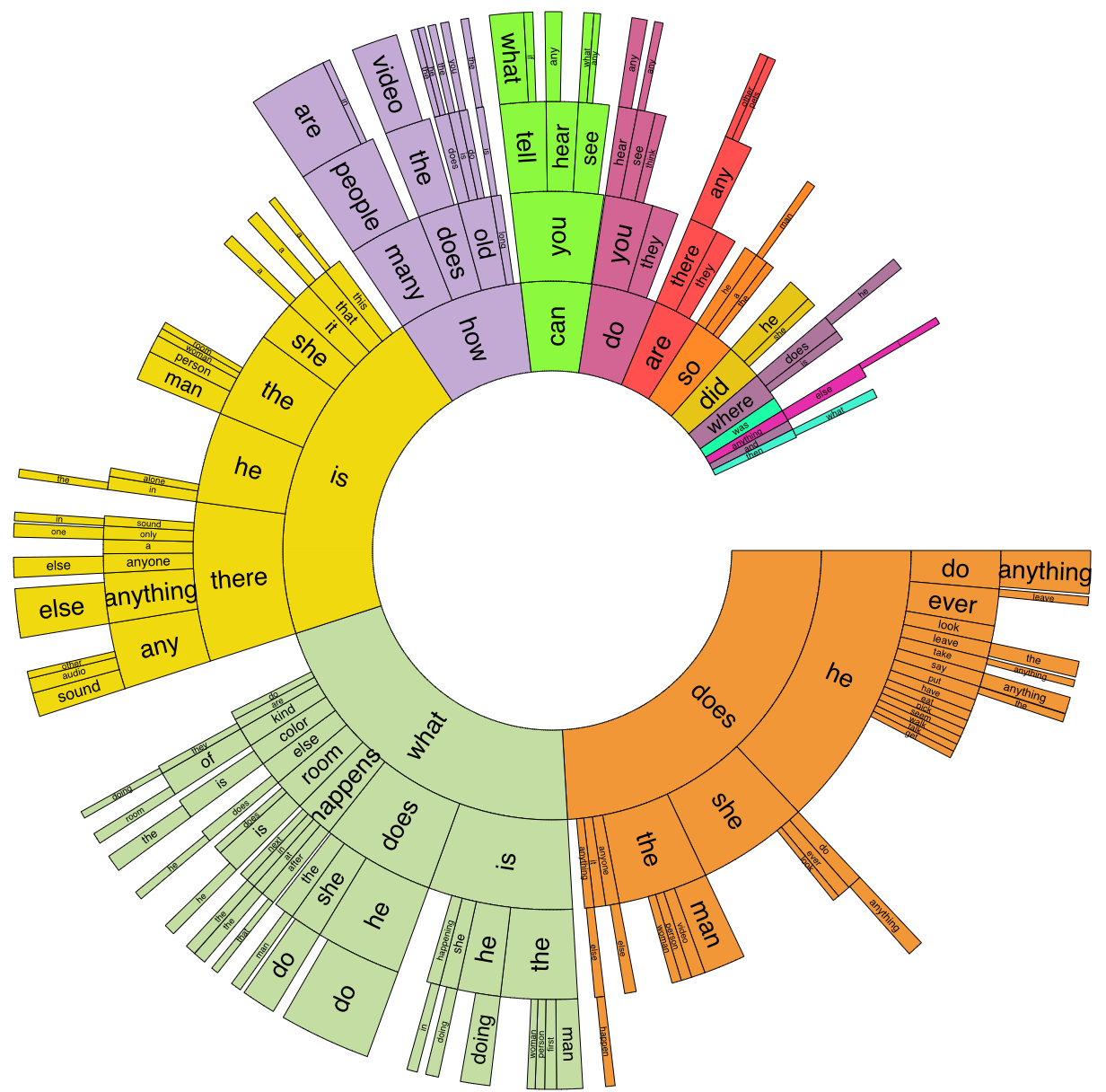}
    \caption{AVSD Questions}
    \label{fig:avsdQue}
    \end{subfigure}
	\begin{subfigure}[t]{0.28\textwidth}
    \includegraphics[width=\linewidth]{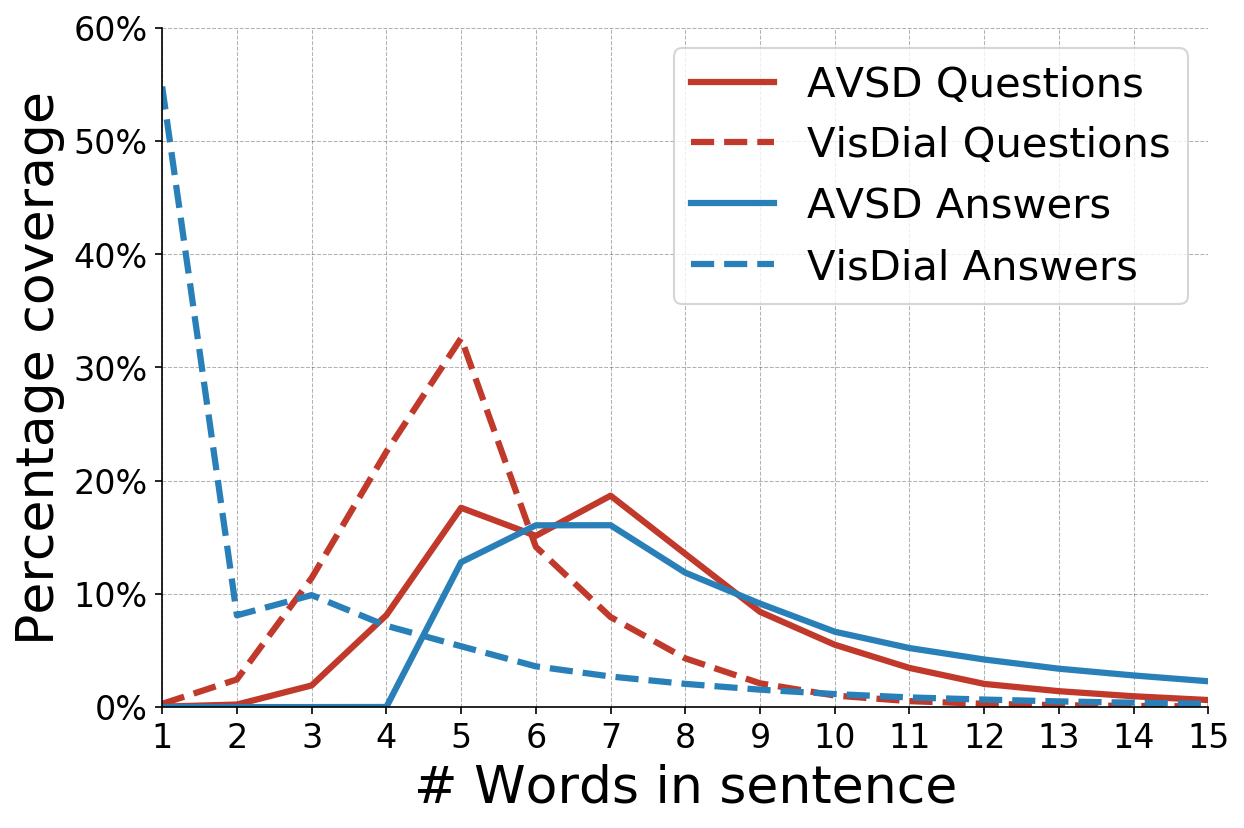}
    \caption{AVSD sentence lengths}
    \label{fig:VisDialAVSD}
    \end{subfigure}
    \caption{Distribution of first n-grams in the AVSD Dataset for (a) AVSD answers and (b) AVSD questions. (c) Distribution of lengths for questions and answers in AVSD compared to those in VisDial \cite{das2017visual}.}
    \vspace{-5pt}
\end{figure*}

In this section, we analyze the new AVSDv1.0 Dataset. In total, the dataset contains 11,816 conversations (7,985 training, 1,863 validation, and 1,968 testing), each including a video summary (written by the Questioner after each dialog). There are a total of 118,160 question/answer (QA) pairs. Figure \ref{fig:conversation} shows an example from our dataset. More examples can be found in the supplementary section. 

\textbf{Lengths of Questions and Answers.} We compare the length of AVSD questions and answers with those from VisDial \cite{das2017visual} in Figure \ref{fig:VisDialAVSD}. Note that the answers and questions in AVSD are longer on average. The average length for AVSD questions and answers is 7.9 and 9.4 words, respectively. In contrast, VisDial questions average 5.1 words and are answered in 2.9 words on average. This shows that the dialogs in our dataset are more verbose and conversational. 

\textbf{Audio-Related Questions.} 
In $57\%$ of the conversations, there are questions about the audio, such as whether there was any music or noise, or whether the people were talking. Here are some examples of these audio-related questions from the dataset:
\begin{center}
\begin{minipage}[l]{0.8\columnwidth}
{\textit{Does she appear to talk to anyone?}}
{\textit{Do you hear any noise in the background?}}
{\textit{Is there any music?}} 
{\textit{Is there any other noise like a TV or music?}}
\end{minipage}
\end{center}

Moreover, looking at the burst diagram for questions in Figure \ref{fig:avsdQue} we can see that questions like ``Can / Do you hear ...'' and ``Is there any sound ...'' appear frequently in the dataset.

\textbf{Temporal Questions.} Another common type of questions is about \textit{what happened next}. In fact, people asked questions about what happened next in more that $70\%$ of the conversations.  As previously noted, the investigation of the temporal sequence of events was implicitly encouraged by our experimental protocol, such as providing the Questioner with three image frames from different parts of the video. Here are some examples of such questions, taken from different conversations:
\vspace{-3pt}
\begin{center}
\begin{minipage}[l]{0.8\columnwidth}
\textit{Does he do anything after he throws the medicine away?}
\textit{Where does she lay the clothes after folding them?}
\textit{What does he do after locking the door?}
\end{minipage}
\end{center}

Likewise, we see that questions such as ``What happens ...'' and ``What does he do ...'' occur frequently in the dataset, as shown in Figure \ref{fig:avsdQue}.\\[-10pt]

\begin{table}[t]
\centering
\setlength{\tabcolsep}{0.3em}
 \resizebox{\columnwidth}{!}{
 \begin{tabular}{l c c c c c c c} 
  & BLEU$_1$ & BLEU$_2$ & BLEU$_3$ & BLEU$_4$ & METEOR & ROUGE$_L$ & CIDEr \\  
 \midrule
video-watcher  & 0.638 & 0.433 & 0.287 & 0.191 & 0.223 & 0.407 & 0.429 \\
Questioner  & 0.560 & 0.379 & 0.249 & 0.165 & 0.191 & 0.369 & 0.297 \\
 \bottomrule
\end{tabular}}
\caption{Comparison on different metrics of a video-watcher summary vs. the 3 other video-watcher summaries, and the Questioner's summary vs. the 3 other video-watcher summaries.}
\vspace{-10pt}
\label{tab:summaries_comparison}
\end{table}

\textbf{Dataset Quality.} In order to further evaluate dialog quality, we ran another study where we asked AMT workers to watch and summarize the videos from the AVSD Dataset. The instruction was ``Summarize what is happening in the video''. 
We collected 4 summaries per video and used the BLEU \cite{papineni2002bleu}, ROUGE \cite{lin2004rouge}, METEOR \cite{banerjee2005meteor} and CIDEr \cite{vedantam2015cider} metrics to compare the summaries collected from the video-watcher to the ones provided by the questioners at the end of each conversation. 
In Table \ref{tab:summaries_comparison}, the first row evaluates a randomly selected video-watcher summary vs.~three others, and the second row evaluates the Questioner's summary vs.~the same three other video-watcher summaries. Both these numbers are close, demonstrating that the Questioners do gain an understanding of the scene from the dialog that is comparable to having watched the video.

 

\section{Model}
\label{sec:model}

To demonstrate the potential and the challenges of this new dataset, we design and analyze a video-dialog answerer model.
The model takes as input a video, the audio track of the video, a dialog history (which comprises the ground-truth script from the Charades dataset and the first $t\!-\!1$ QA pairs of the dialog), and a follow-up question (the $t$th question in the dialog). The model should ground the question in both the video and its audio, and use the dialog history to leverage contextual information in order to answer. 

Moving away from the hierarchical or memory network encoders common for dialog tasks \cite{das2017visual}, we opt to present a straightforward, discriminative late-fusion approach for scene-aware dialog that was recently shown to be effective for visual dialog \cite{jain2018two}. This choice also enables a fair ablation study for the various input modalities, an important endeavour when introducing such a strongly multimodal task. For this class of model architecture, increases or decreases in performance from input ablation are directly linked to the usefulness of the input rather than to any complications introduced by the choice of network structure (e.g.,~some modalities having many more parameters than others).

\begin{figure*}[t]
    \centering
    \includegraphics[width=0.78\textwidth]{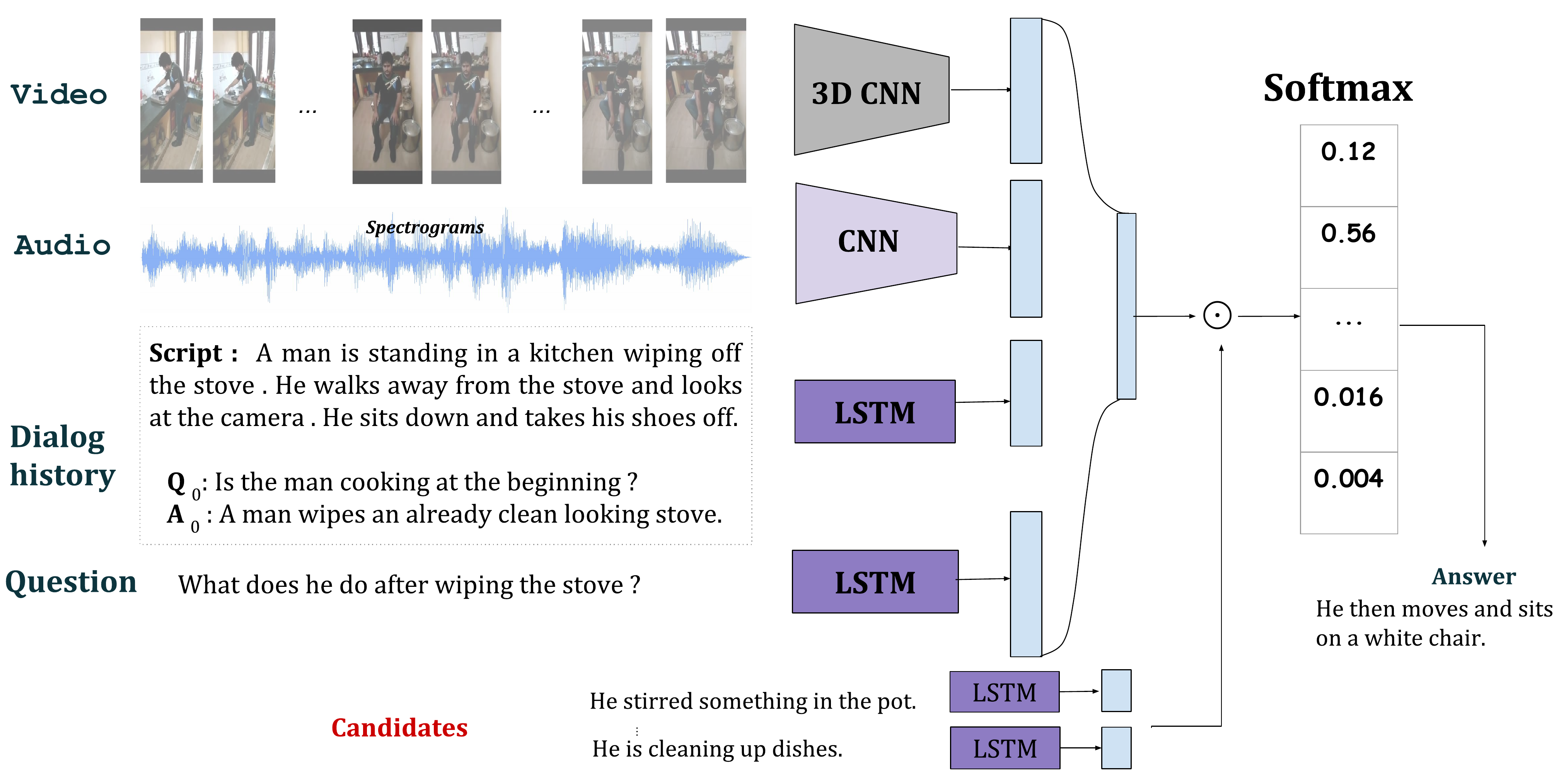}
   \caption{An overview of our late-fusion multimodal network. The encoder takes each input modality and transforms them to a state embedding that is used to rank candidate answers.} 
   \vspace{-10pt}
\label{fig:network}
\end{figure*}

An overview of our model 
is shown in Figure~\ref{fig:network}. At a high level, the network operates by fusing information from all of the modalities into a fixed-size representation, then comparing this state with a set of candidate answers, selecting the most closely matching candidate as the output answer. In the rest of this section, we provide more details of the model and the input encodings for each modality.

\xhdr{Input Representations.} The AVSD Dataset provides a challenging multimodal reasoning task including natural language, video, and audio. We describe how we represent each of these as inputs to the network. These correspond to the information that was available to the human Answerer in round $t$ of a dialog.
\vspace{3pt}
\begin{compactitem}
\item \textbf{Video Script (S)}: Each dialog in AVSD starts with a short natural language description of the video contents (i.e., the Charades ground-truth script).
\item \textbf{Dialog History (DH):} The dialog history consists of the initial video script (S) and each of the question-answer pairs from previous rounds of dialog. At round $t$, we write the dialog history as $\mbox{DH}_t{=}(S, Q_0, A_0, Q_1, A_1, \dots Q_{t-1}, A_{t-1})$. We concatenate the elements of the dialog history and encode them using an LSTM trained along with the late-fusion model.
\item \textbf{Question (Q):} The question to be answered, also known as $Q_{t}$. The question is encoded by an LSTM trained along with the late-fusion model.
\item \textbf{Middle Frame (I):} In some ablations, we represent videos using only their middle frame to eliminate all temporal information, in order to evaluate the role of temporal visual reasoning. In these cases, we encode the frame using a pretrained VGG-16 network \cite{vgg} trained on ImageNet~\cite{imagenet}.
\item \textbf{Video (V):} Each AVSD dialog is grounded in a video that depicts people performing simple actions. We transform the video frames into a fixed sized feature using the popular pretrained I3D model~\cite{carreira2017quo}. I3D is a 3D convolutional network that achieved state-of-the-art performance on multiple popular activity recognition tasks \cite{kuehne2011hmdb, soomro2012ucf101}. 
\item \textbf{Audio (A):} We similarly encode the audio track from the video using a pretrained AENet  \cite{takahashi2018aenet}. AENet is a convolutional audio encoding network that operates over long-time-span spectrograms. It has been shown to improve activity recognition when combined with video features.
\end{compactitem}

\xhdr{Encoder Network.} In order to combine the features from these diverse inputs, we follow recent work in visually grounded dialog \cite{jain2018two}: simply concatenate the features, and allow fusion to occur through fully-connected layers. More concretely, we can write our network as:

\vspace{-5pt}
\begin{equation*}\label{eq:eq_lstm}
\begin{aligned}
h_t &= \mathrm{LSTM}(\mathrm{DH}) \quad &
q_t &= \mathrm{LSTM}(\mathrm{Q}) \\
i &= \mathrm{I3D}(\mathrm{V}) \quad &
a &= \mathrm{AENet(A)} \\
\end{aligned}
\end{equation*}
\begin{equation*}\label{eq:eq_late_fusion}
\begin{aligned}
z &= \mathrm{concat}(h_t, q_t, i, a) \\
\vspace{-3pt}
e_n &= \mathrm{tanh}\left( \textstyle{\sum_{k}} \,w_{k,n} \times z_k + b_n \right), \\
\end{aligned}
\end{equation*}
\noindent where $h_t$, $q_t$, $i$, and $a$ are the dialog history, question, video, and audio feature embeddings  described above. The embeddings are concatenated to form the vector $z$, which is passed through a linear layer with a tanh activation to form the joint embedding vector $e$. (Here $k$ and $n$  respectively index elements of the vectors $z$ and $e$.) For any of our ablations of these input modalities, we simply train a network excluding that input, without adjusting the linear layer output size.

\xhdr{Decoder Model.} 
We approach this problem as a discriminative ranking task, selecting an output from a set of candidate options, since these approaches have proven to be stronger than their generative counterparts in visual dialog \cite{das2017visual}. (However, we note that generative variants need not rely on a fixed answer pool and may be more useful in general deployment.) More concretely, given a set of 100 potential answers $\{\mathcal{A}_t^{(1)}, \dots, \mathcal{A}_t^{(100)}\}$, the agent learns to pick the most appropriate response.

The decoder computes the inner product between a candidate answer embedded with an LSTM and the holistic input embedding $e$ generated by the encoder.  We can write the decoder as:
%
\begin{equation}\label{eq:eq_classif}
\begin{aligned}
a_{t,i} &= \mathrm{LSTM}(\mathcal{A}_t^{(i)})\\
s_{t,i} &= \;<a_{t,i}, e>
\end{aligned}
\end{equation}
\vspace{2pt}
\noindent where $a_{t,i}$ is the embedding vector for answer candidate $\mathcal{A}_t^{(i)}$,  the notation $<\!\cdot, \cdot\!>$ represents an inner product, and $s_{t,i}$ is the score computed for the candidate based on its similarity to the input encoding $e$. We repeat this for all of the candidate answers, then pass the results through a softmax layer to compute probabilities of all of the candidates. At training time, we maximize the log-likelihood of the correct answer. At test time, we simply rank candidates according to their probabilities and select the argmax as the best response. 

\label{sec:cand}
\textbf{Selecting Candidate Answers.}
Following the selection process in \cite{das2017visual}, the set of 100 candidates answers consists of four types of answers: the ground-truth answer, hard negatives that are ground-truth answers to similar questions (but different video contexts), popular answers, and answers to random questions. We first sample 50 plausible answers which are the ground-truth answers to the 50 most similar questions. We are looking for questions that start with similar tri-grams (i.e.,~are of the same type such as ``what did he'') and mention similar semantic concepts in the rest of the question. To accomplish this, all the questions are embedded in a common vector space. The question embedding is computed by concatenating the GloVe \cite{pennington2014glove} embeddings of the first three words with the averaged GloVe embedding of the remaining words in the question. We then use Euclidean distance to select the closest neighbor questions to the original question. Those sampled answers are considered as hard negatives, because they correspond to similar questions that were asked in completely different contexts (different video, audio and dialog). 
In addition, we select the 30 most popular answers from the dataset. By adding popular answers, we force the network to distinguish between purely likely answers and plausible responses for the specific question, which increases the difficulty of the task.
The next 19 candidate answers are sampled from the ground-truth answers to random questions in the dataset. The final candidate answer is the ground-truth (human-generated) answer from the original dialog.

\textbf{Implementation Details.}
Our implementation is based on the visual dialog challenge starter code \cite{githubvisdialchalstarterpytorch}. The VisDial repository also provides code and model to extract image features. We extract video features using the I3D model \cite{carreira2017quo}. Repository \cite{githubi3d} provides code and models fine-tuned on the Charades dataset to extract I3D video features. We subsample 40 frames from the original video and feed them into the RGB pipeline of the I3D model. The frames are sampled to be equally spaced in time. For the audio features, we use the AEnet network \cite{takahashi2018aenet}. The repository \cite{githubaenet} provides code to extract features from an audio signal. We first extract the audio track from the original Charades videos and convert them into 16kHz, 16bit, mono-channel signals. Both the video and audio features have the same dimension (4096).

\vspace{\sectionReduceTop}
\section{Experiments}
\label{sec:exp}

\xhdr{Data Splits.} Recall from Section \ref{sec:dataset} that the AVSDv1.0 dataset contains 11,816 instances split across training (7,985), validation (1,863), and testing (1,968) corresponding to the source Charades video splits. We present results on the test set.

\xhdr{Evaluation Metrics.}
Although metrics like BLEU \cite{papineni2002bleu}, METEOR \cite{banerjee2005meteor}, and ROUGE \cite{lin2004rouge} have been widely used to evaluate dialog \cite{lowe2015ubuntu,sordoni2015neural,zhou2018dataset}, there has been recent evidence suggesting that they do not correlate well with human judgment \cite{liu2016not}. Like \cite{das2017visual},  we instead evaluate our models by checking individual responses at each round in a retrieval or multiple-choice setting. The agent is given a set of 100 answer candidates (Section \ref{sec:cand}) and must select one. We report the following retrieval metrics:
\begin{compactitem}
    \item \textbf{Recall@k [higher is better]} that measures how often the ground truth is ranked in the top $k$ choices
    \item \textbf{Mean rank (Mean) [lower is better]} of the ground truth answer which is sensitive to overall tendencies to rank ground-truth higher---important in our context as other candidate answers may be equally plausible.
    \item \textbf{Mean reciprocal rank (MRR) [higher is better]} of the ground truth answer, which values placing ground truth in higher ranks more heavily.
\end{compactitem}
We note that evaluation even in these retrieval settings for dialog has many open questions. One attractive alternative that we leave for future work is to evaluate directly with human users in cooperative tasks \cite{visdial_eval_hcomp_2017}.

\begin{figure*}[ht]
\begin{center}
\resizebox{\textwidth}{!}{
	\begin{subfigure}[b]{0.45\textwidth}
	    \centering
	    \includegraphics[width=\linewidth]{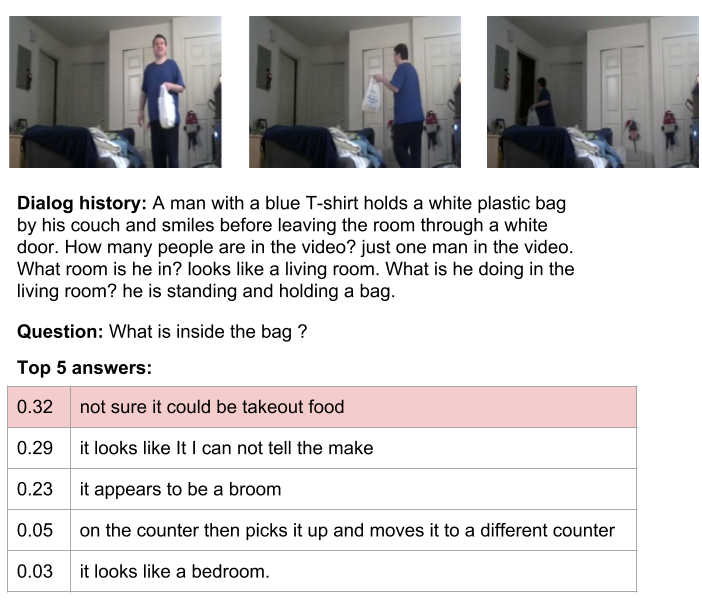}
	\end{subfigure}
	\hspace{10pt}
	\vrule
	\hspace{10pt}
	\begin{subfigure}[b]{0.45\textwidth}
	    \centering
	    \includegraphics[width=\linewidth]{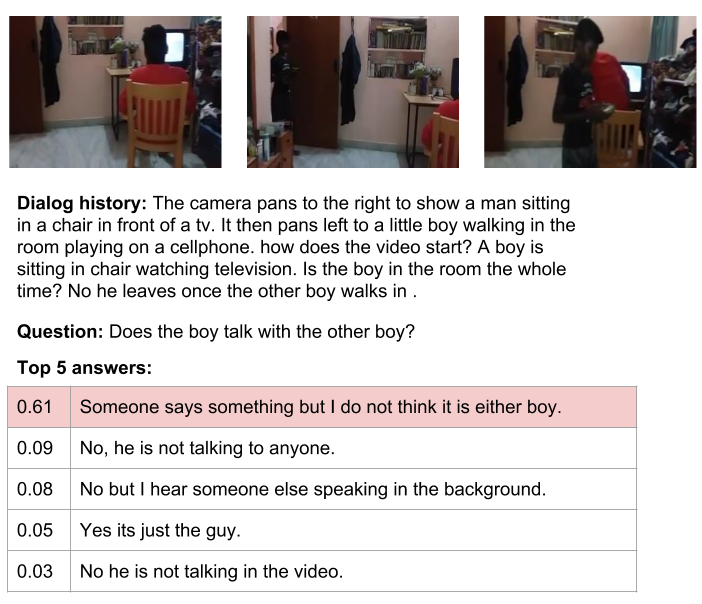}
	\end{subfigure}
	}
\end{center}
    \vspace{-10pt}
	\caption{Example using \texttt{Q+DH+V+A}. The left column of the tables in each figure represents the corresponding answer probability. The ground truth answer is highlighted in red. In both of these examples, the model ranked the ground truth answer at top position.}
    \label{fig:samples}
\end{figure*}

\vspace{\sectionReduceTop}
\section{Results and Analysis}

In order to assess the challenges presented by the AVSDv1.0 dataset and the usefulness of different input modalities to address them, we present comprehensive ablations of our baseline model with respect to inputs. Table \ref{tab:results} reports the results of our models on AVSDv1.0 test. We find that our best performing models are those that can leverage video, audio, and dialog histories---signaling that the dialog collected in AVSD is grounded in multi-modal observations. In the rest of this section, we highlight noteworthy results.

\begin{table}[b]
\centering
\renewcommand*{\arraystretch}{1.15}
\resizebox{\columnwidth}{!}{
 \begin{tabular}{c c c c c c c} 
 \toprule
& Model & MRR & R@1 & R@5 & R@10 & Mean\\  
 \midrule
\multirow{4}{*}{\rotatebox{90}{\shortstack{Language\\Only}}}& Answer Prior & 7.85 & 1.66 & 8.17 & 16.54 & 28.54 \\
& Q & 36.12 & 20.01 & 53.72 & 74.55 & 7.63 \\\
& Q + C& 37.42 & 20.95& 56.17 & 76.60 & 7.27\\
& Q + DH & 50.40 & 32.76 & 73.27 & 88.60 & 4.72 \\
\midrule
\multirow{4}{*}{\rotatebox{90}{\shortstack{Perception w/o \\ Dialog Context}}}& Q + I & 35.12 & 19.08 & 52.36 & 73.35 & 7.90 \\ 
& Q + V & 39.36 & 22.32 & 59.34 & 78.65 & 6.86 \\
& Q + A & 35.94 & 19.46 & 54.55 & 75.14 & 7.58 \\
& Q + V + A & 38.83 & 22.02 & 58.17 & 78.18 & 7.00 \\[-6pt]
\\
\midrule
\multirow{3}{*}{\rotatebox{90}{\shortstack{Full \\ Models}}}& Q + DH + I & 50.52 & 32.98 & 73.26 & 88.39 & 4.73 \\
& Q + DH + V  & \textbf{53.41} & \textbf{36.22} & \textbf{75.86} & 89.79 & 4.41 \\
&Q + DH + V + A & 53.03 & 35.65 & 75.76 & \textbf{89.92} & \textbf{4.39} \\
 \bottomrule
 \label{table:results}
\end{tabular}}\vspace{-8pt}
\caption{Results of model ablations on the AVSDv1.0 test split. We report mean receiprocal rank (MRR), recall@k (R@K), and mean rank (Mean). We find that our best performing model leverages the dialog, video, and audio signals to answer questions.}
\label{tab:results}
\end{table}

\xhdr{Language-only Baselines.} The first four lines of Table \ref{tab:results} show the language-only models. First, the \texttt{Answer Prior} model encodes each answer with an LSTM and scores it against a static embedding vector learned over the entire training set. This model lacks question information, caption, dialog history, or any form of perception, and acts as a measure of dataset answer bias. Naturally, it performs poorly over all metrics, though it does outperform chance.  We also examine a question-only model \texttt{Q} that selects answers based only on the question encoding, a question and a caption model \texttt{Q+C}, as well as a question and dialog history \texttt{Q+DH} model that also includes the caption. These models measure regularities between questions, dialogs, and answer distributions. We find that access to the question greatly improves performance over the answer prior from 28.54 mean rank to 7.63 with question alone. While caption encoding has no significant impact on the model performance, adding the dialog history provides the best language-only model performance of 4.72 mean rank.

\xhdr{Dialog history is a strong signal.} The dialog history appears to be a very strong signal -- models with it consistently achieve mean ranks in the 4--4.8 range even without additional perception modalities, whereas models without dialog history struggle to get below a mean rank of 7. This makes sense, as dialogs are self-referential; in the AVSD dataset, 55.2\% of the questions contain co-reference words such as {\em her}, {\em they}, and {\em it}. Such questions strongly depend on the prior rounds of dialog, which are encoded in the DH.
We note that adding video and audio signals improves over dialog history alone, by providing complementary information to ground questions.

\xhdr{Temporal perception seems to matter.} Adding video features (\texttt{V}) consistently leads to improvements for all models. To further tease apart the effect of temporal perception from being able to see the scene in general, we run two ablations where rather than the video features, we encode visual perception using only the middle frame of the video. In both cases, \texttt{Q+I} and \texttt{Q+DH+I}, we see that the addition of static frames hurts performance marginally whereas addition of video features leads to improvements. Thus, it seems that whereas temporal perception is helpful, models with access to just the middle image learn poorly generalizable groundings. We point out that one confounding factor for this finding is that the image is encoded with a VGG network, rather than the I3D encoding used for videos.

\xhdr{Audio provides a boost.} The addition of audio features generally improves model performance (\texttt{Q+V} to \texttt{Q+V+A} being the exception). Interestingly, we see that model performance 
improves 
even more when combined with dialog history and video features (\texttt{Q+DH+V+A}) for some metrics, indicating there is still complementary knowledge between the video and audio signals despite their close relationship.

\xhdr{Temporal and Audio-Based Questions.} Table \ref{tab:audio_vs_visual_results} shows mean rank on subsets of questions. We filter the questions using the two lists of keywords: audio-related words \{talk, hear, sound, audio, music, noise\} and
temporal words \{after, before, beginning, then, end, start\}.
We then generated answers to those questions using the three different models \texttt{Q}, \texttt{Q+A} and \texttt{Q+V} and compared which one would lead to higher rank of the ground truth answer. \\[-10pt]

\begin{table}[t]
\centering
\renewcommand*{\arraystretch}{1.0}
\resizebox{\columnwidth}{!}{
\begin{tabular}{c c c c c c c}
 &Model & MRR & R@1 & R@5 & R@10 & Mean\\
 \bottomrule 
 \multirow{3}{*}{\rotatebox{90}{\shortstack{Original\\ Setting}}} & Q + DH & 50.40 & 32.76 & 73.27 & 88.60 & 4.72 \\
 & Q + DH + V & 53.41 & 36.22 & 75.86 & 89.79 & 4.41\\
 & Q + DH + V + A & 53.03 & 35.65 & 75.76 & 89.92 & 4.39\\
 \bottomrule
 \multirow{3}{*}{\rotatebox{90}{\shortstack{Shuffled}}} &
 Q + DH  & 49.03 & 31.55 & 71.28 & 86.90 & 5.03 \\
 &Q + DH + V& 51.47 & 34.17 & 74.03 & 88.40 & 4.72\\
 &Q + DH + V + A& 50.74 & 33.22 & 73.20 & 88.27 & 4.76\\
\bottomrule
 \end{tabular}}
 \caption{Shuffling the order of Questions (Q/A pairs). \textit{Original Settings:} Original results. \textit{Shuffled:~}  Results on shuffled dialogs.} 
 \label{tab:resutls}
 \vspace{-15pt}
 \end{table}

\begin{table}[h]
\centering
\renewcommand*{\arraystretch}{1.15}
 \begin{tabular}{c c c c} 
  & \texttt{Q} & \texttt{Q+A} & \texttt{Q+V} \\
 \midrule
\multirow{1}{*}
Audio questions & 6.91 & 6.69 & \textbf{6.52}  \\
Temporal questions & 7.31 & 7.15 & \textbf{5.98} \\
 \bottomrule
\end{tabular}
\caption{Mean rank results for the three models \texttt{Q}, \texttt{Q+A}, and \texttt{Q+V} for audio-related questions and temporal questions.}
\label{tab:audio_vs_visual_results}
\end{table}

For the audio-related questions, we can see that although both the \texttt{Q+A} and \texttt{Q+V} outperform the \texttt{Q} model, the visual features seem more useful. This can be  easily balanced as it is also unlikely that vision is unnecessary in audio questions. However, answers to the temporal questions were much better using the \texttt{Q+V} model, which confirms our intuition. The \texttt{Q+A} model helps only slightly ($7.15$ vs $7.31$), but the \texttt{Q+V} model yields more significant improvement ($5.98$ vs $7.31$).

\xhdr{The order of the questions/answers is important.}
An important question to ask is whether the questions and the answers in the dialog are a set of independent question/answer (QA) pairs, or are they strongly co-dependent? To answer this question, we ran an experiment in which we tested the trained model on a shuffled test set containing randomly ordered QA pairs. The top section of Table \ref{tab:resutls} shows the results on the original test set (ordered), with the results on the shuffled test set below. We observe a difference of $\sim\!\! 1.87$ for R@$k$ averaged across $k$ and models, and $\sim \!\!0.33$ for the mean rank averaged across models, indicating that the order of the QA pairs indeed matters.

\xhdr{Qualitative Examples.} Figure \ref{fig:samples} shows two examples using the setup \texttt{Q+DH+V}. The first column in the answer table of each example is the answer probability. The ground truth answer is highlighted in red.


\vspace{\sectionReduceTop}
\section{Conclusion}
\vspace{\sectionReduceBot}
\label{sec:conclusions}
We introduce a new AI task: Audio Visual Scene-Aware Dialog, where the goal is to hold a dialog by answering a user's questions about dynamic scenes using natural language. We collected the Audio Visual Scene-Aware Dialog (AVSD) Dataset, using a two-person chat protocol on more than 11,000 videos of human actions. We also developed a model and performed many ablation studies, highlighting the quality and complexity of the data. Our results show that the dataset is rich, with all of the different modalities of the data playing a role in tackling this task. We believe our dataset can serve as a useful benchmark for evaluating and promoting progress in audiovisual intelligent agents. 

\newpage

{
\small
\bibliographystyle{ieee_fullname}
\bibliography{egbib}
}
\clearpage
\section*{Appendix Overview}
This supplementary document is organized as follow:
\begin{itemize}
    \item Sec. \ref{sec:avsd_examples} Dialog examples from AVSD.
    \item Sec. \ref{sec:section_qualitative_exampels} qualitative examples from AVSD. 
    \item Sec. \ref{sec:section_summaries_interface} snapshots of our Amazon Mechanical Turk interface that served collecting the video summaries along with some examples.
\end{itemize}

\addcontentsline{toc}{section}{Appendices}
\renewcommand{\thesection}{\Alph{section}}

\vspace{\sectionReduceTop}
\section{AVSD Examples}
\label{sec:avsd_examples}

\begin{figure*}[ht]
\begin{center}
	\begin{subfigure}[b]{0.45\textwidth}
	    \centering
	    \includegraphics[width=\linewidth]{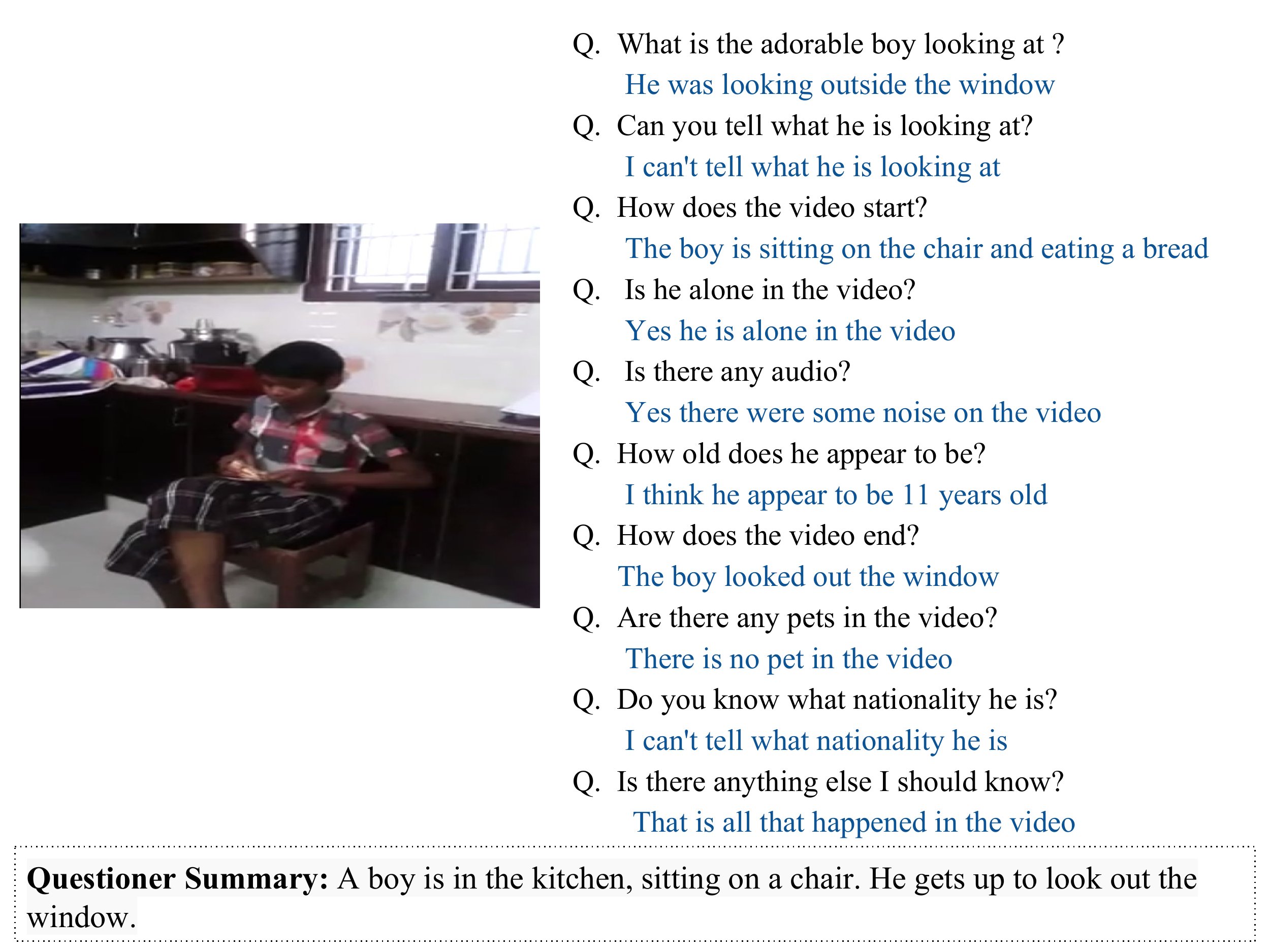}
	    \caption{ }
	    \label{fig:avsd_1}
	\end{subfigure}
	\begin{subfigure}[b]{0.45\textwidth}
	    \centering
	    \includegraphics[width=\linewidth]{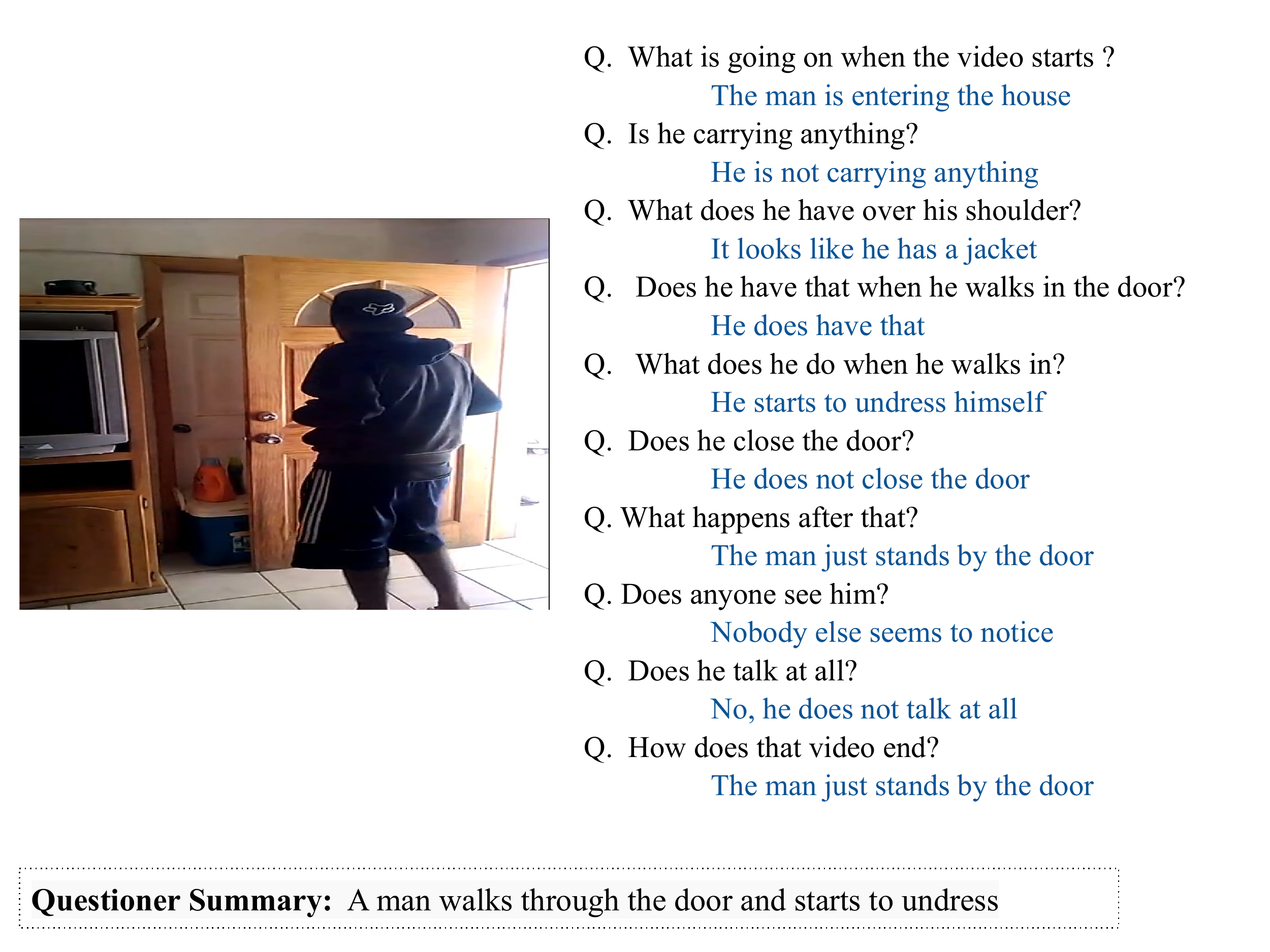}
	    \caption{ }
	    \label{fig:avsd_2}
	\end{subfigure}
	
	\begin{subfigure}[b]{0.45\textwidth}
	    \centering
	    \includegraphics[width=\linewidth]{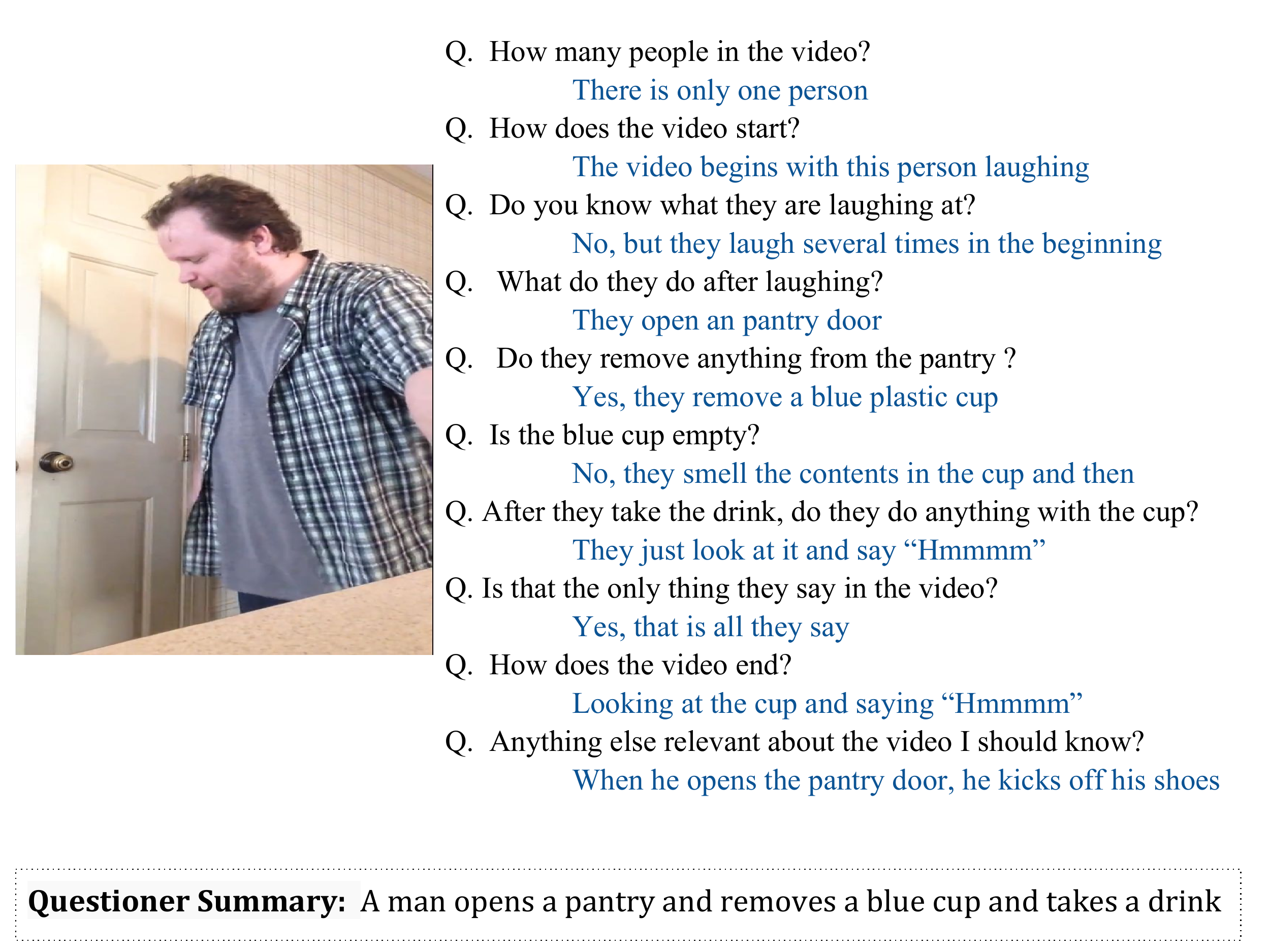}
	    \caption{ }
	    \label{fig:avsd_3}
	\end{subfigure}
	\begin{subfigure}[b]{0.45\textwidth}
	    \centering
	    \includegraphics[width=\linewidth]{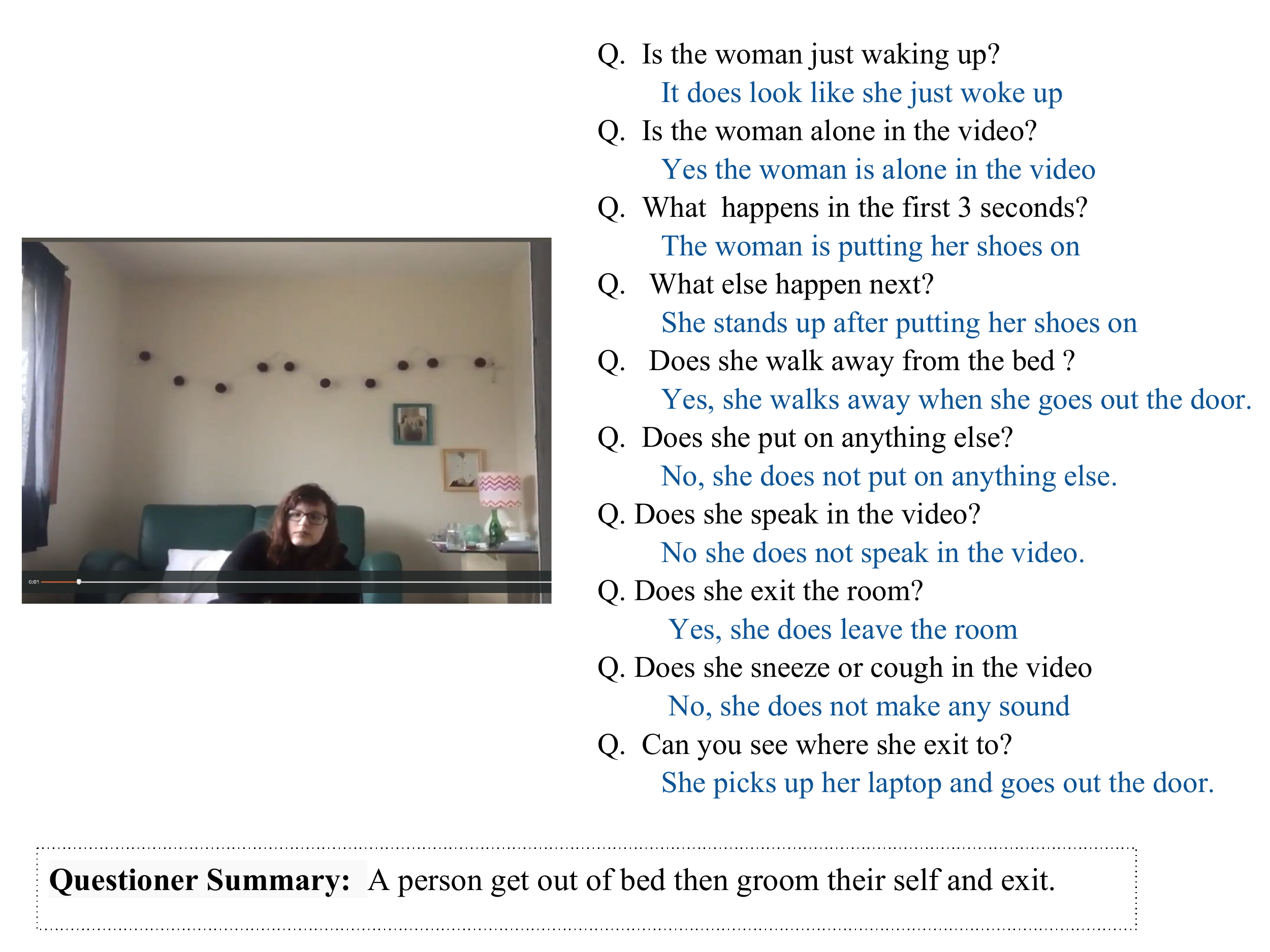}
	    \caption{ }
	    \label{fig:avsd_4}
	\end{subfigure}
	
	\begin{subfigure}[b]{0.45\textwidth}
	    \centering
	    \includegraphics[width=\linewidth]{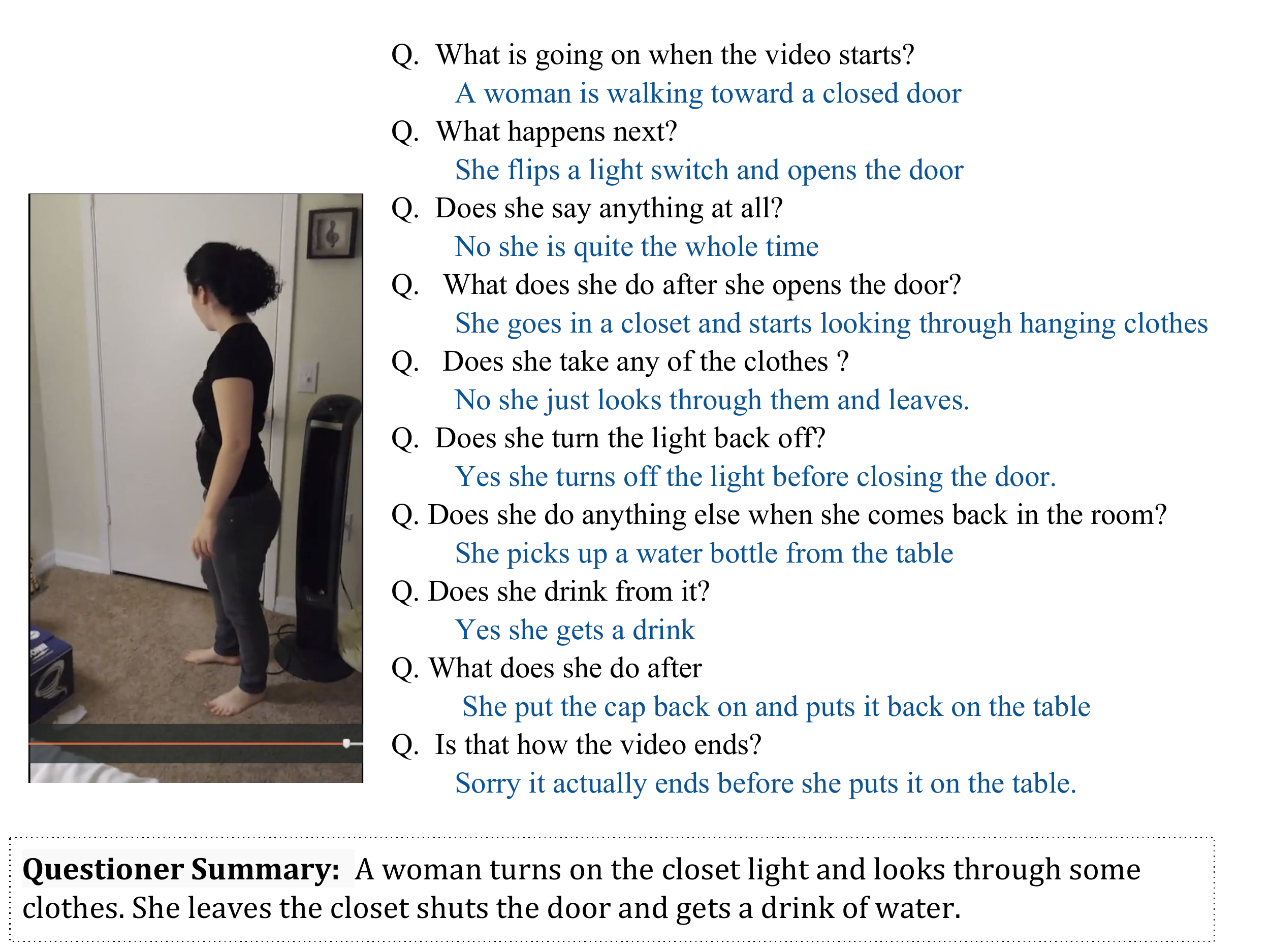}
	    \caption{ }
	    \label{fig:avsd_5}
	\end{subfigure}
	\begin{subfigure}[b]{0.45\textwidth}
	    \centering
	    \includegraphics[width=\linewidth]{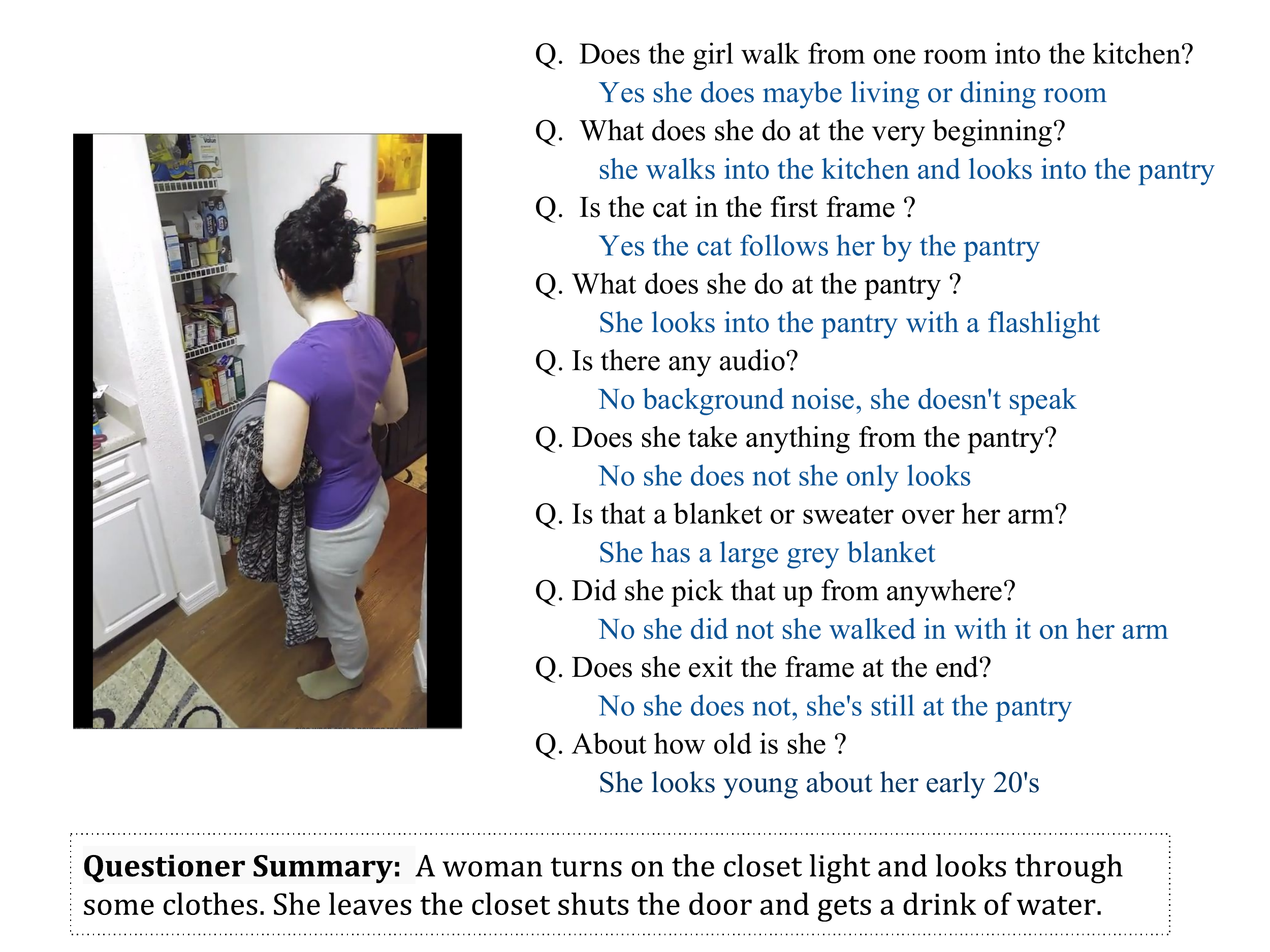}
	    \caption{ }
	    \label{fig:avsd_6}
	\end{subfigure}
\end{center}
	\caption{Examples of AVSD V.1}
    \label{fig:avsd_examples}
\end{figure*}

Examples of AVSD datasets are presented in \ref{fig:avsd_examples}. As we can see from these examples, the questions cover several aspects of the video. While most of questions ask about the actions in the video, some questions focus on actions at specific temporal segment (e.g.  ``what is happening at the end? ``  ``How does the video start? ``  ``What is happening next? ``. In addition, the dataset also has audio related questions:  ``Is there any audio? ``.  ``Does he talk at all `` ?  ``Is that the only thing they say in the video ``. 

\vspace{\sectionReduceTop}
\section{Qualitative examples from our dataset.}
\label{sec:section_qualitative_exampels}
In this section we discuss the model responses to several types of challenging and interesting questions in our dataset. In a video-based dialog, questions can be about audio, visual appearance, temporal information or actions. We examine the model responses for these questions based on different input modalities. Examples are randomly selected from the test set.

\subsection{Examples with \texttt{Q+DH+V+A}}
\begin{figure*}[ht]
\centering
	\begin{subfigure}[b]{0.45\textwidth}
	    \centering
	    \includegraphics[width=\linewidth]{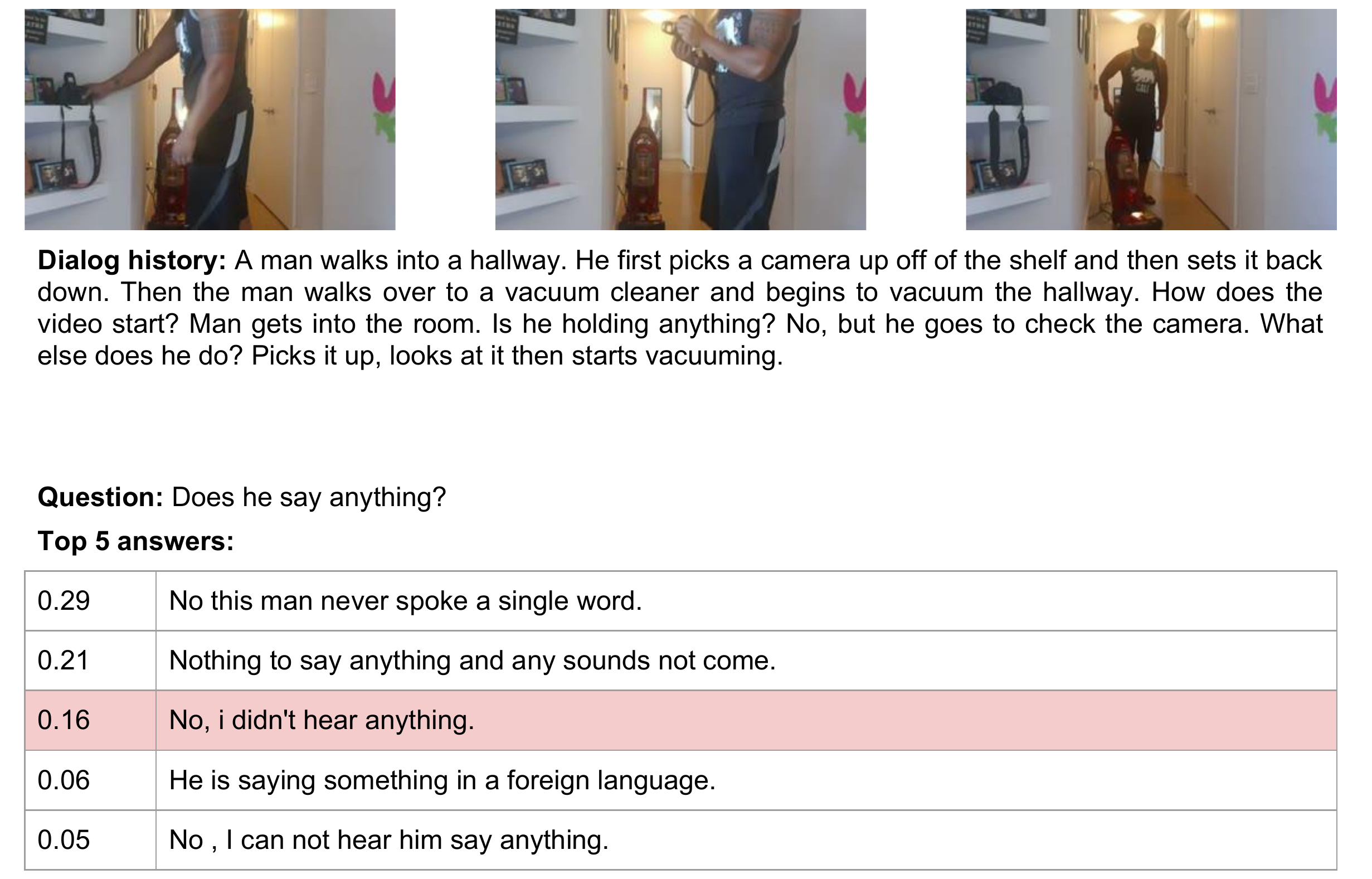}
	    \caption{ }
	    \label{fig:s_a}
	\end{subfigure}
	\hspace{10pt}
	\vrule
	\hspace{10pt}
	\begin{subfigure}[b]{0.45\textwidth}
	    \centering
	    \includegraphics[width=\linewidth]{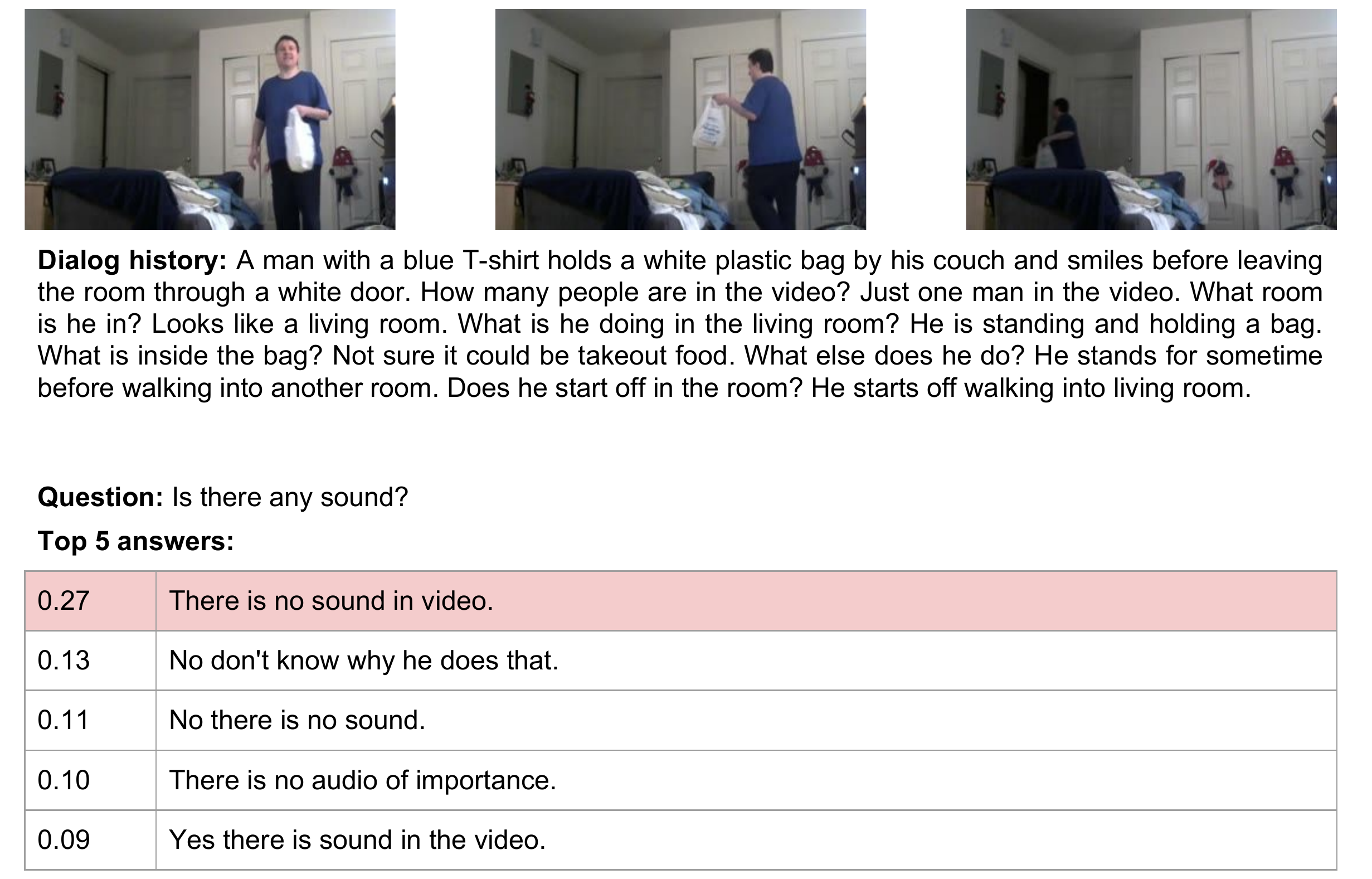}
	    \caption{ }
	    \label{fig:s_b}
	\end{subfigure}%
	
    \vspace{6pt}
	\begin{subfigure}[b]{0.45\textwidth}
	    \centering
	    \includegraphics[width=\linewidth]{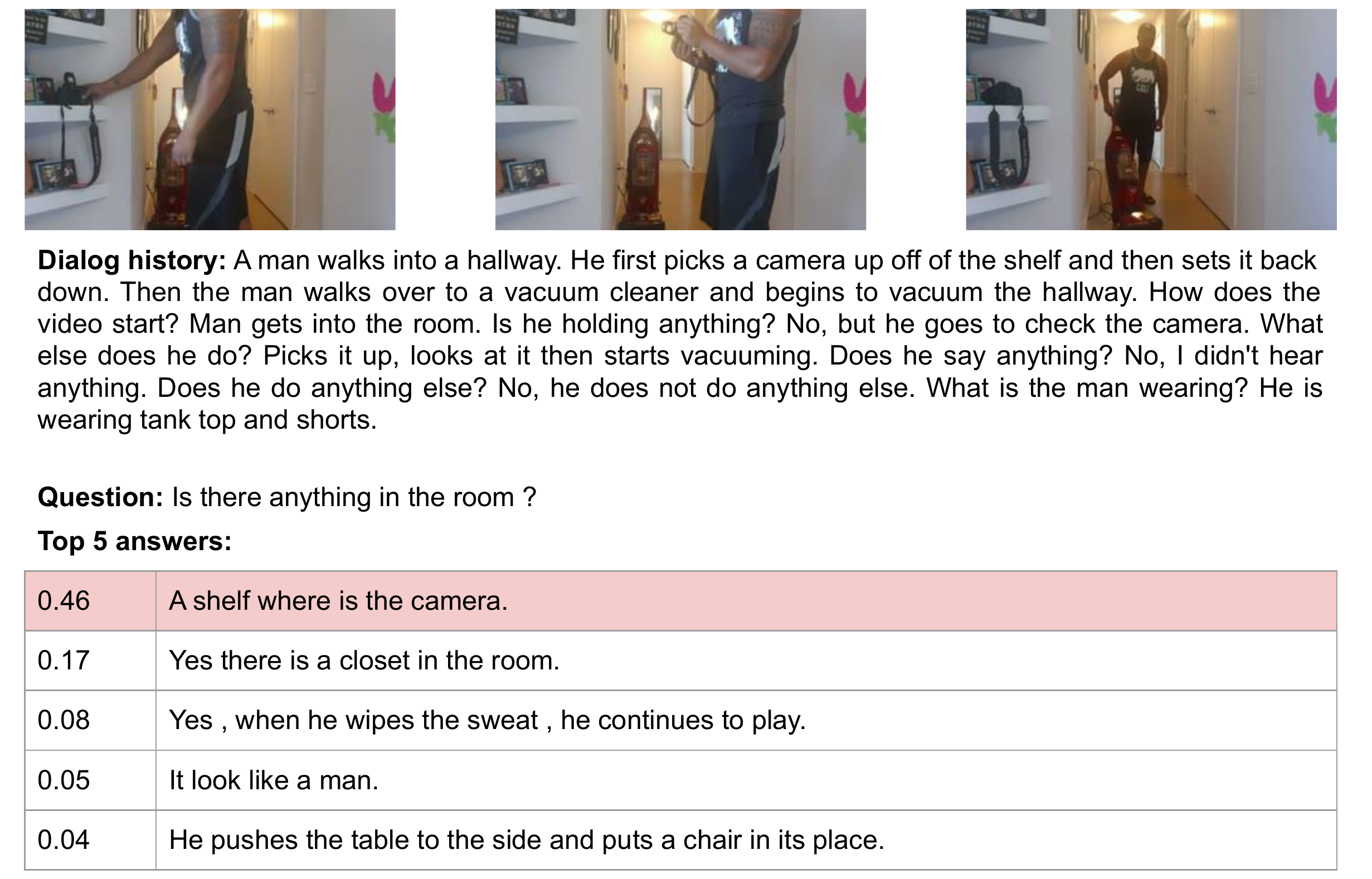}
	    \caption{ }
	    \label{fig:s_c}
	\end{subfigure}
		\hspace{10pt}
	\vrule
	\hspace{10pt}
	\begin{subfigure}[b]{0.45\textwidth}
	    \centering
	    \includegraphics[width=\linewidth]{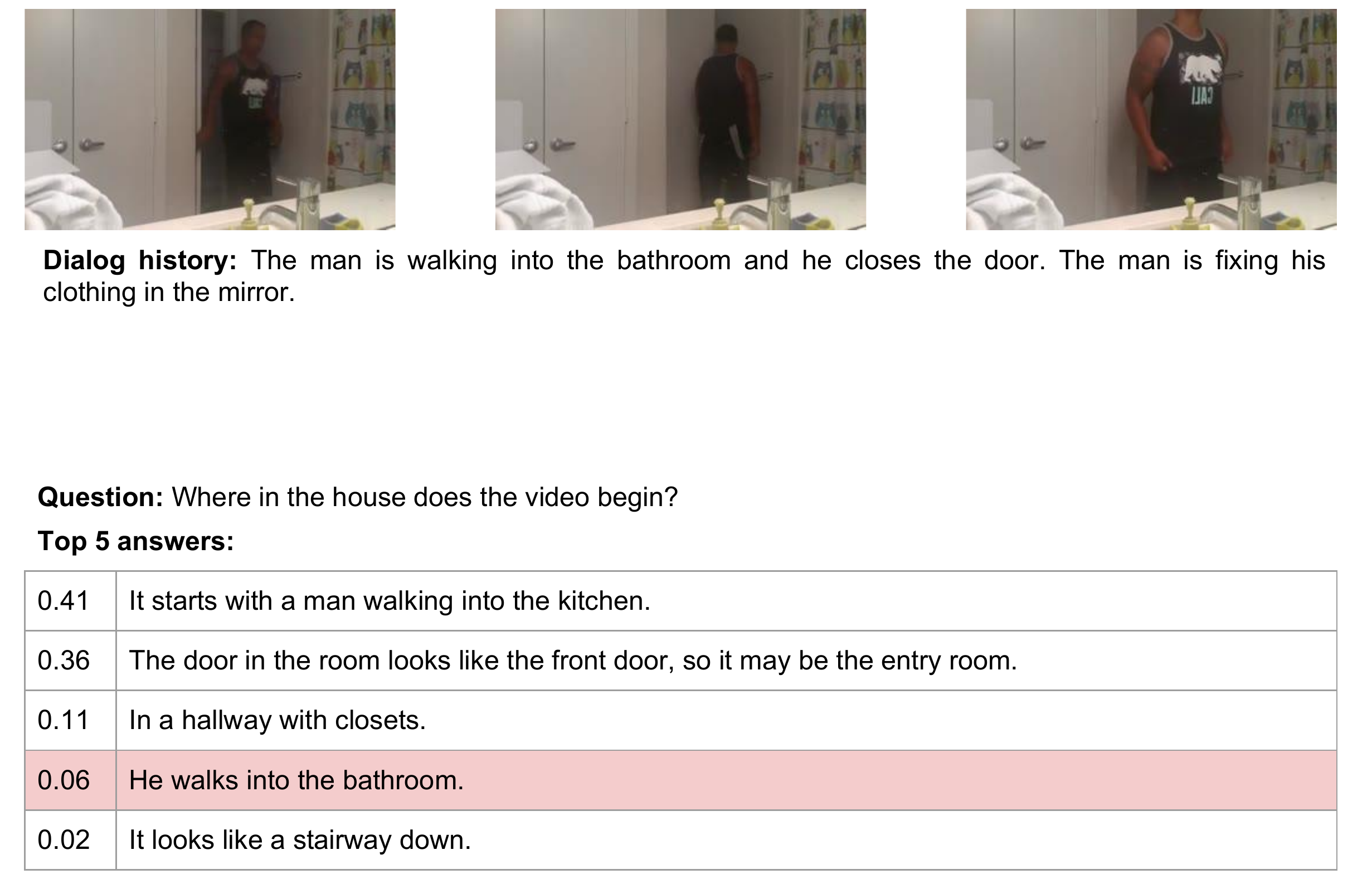}
	    \caption{ }
	    \label{fig:s_d}
	\end{subfigure}
	
    \vspace{6pt}
	\begin{subfigure}[b]{0.45\textwidth}
	    \centering
	    \includegraphics[width=\linewidth]{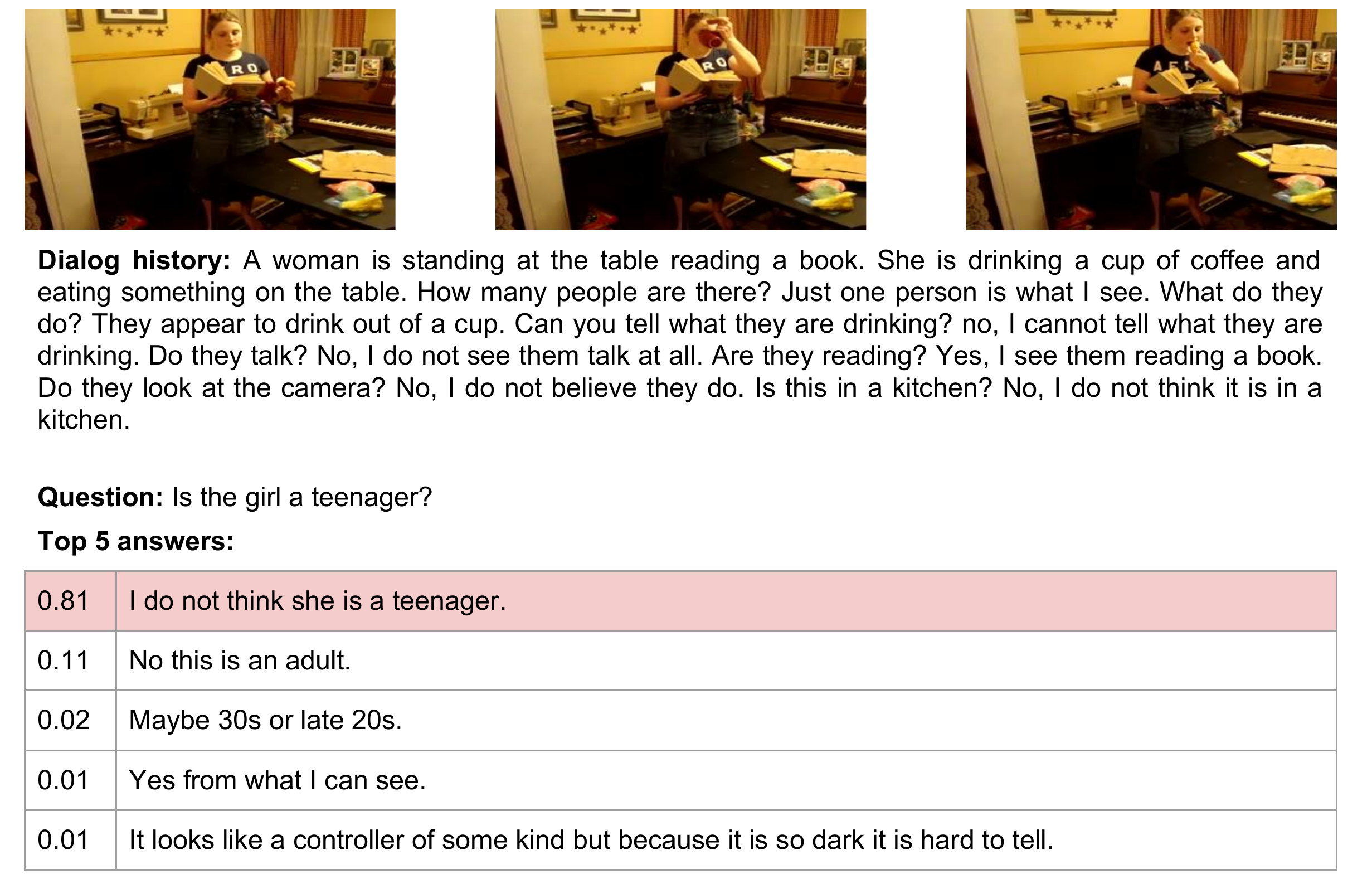}
	    \caption{ }
	    \label{fig:s_e}
	\end{subfigure}
		\hspace{10pt}
	\vrule
	\hspace{10pt}
	\begin{subfigure}[b]{0.45\textwidth}
	    \centering
	    \includegraphics[width=\linewidth]{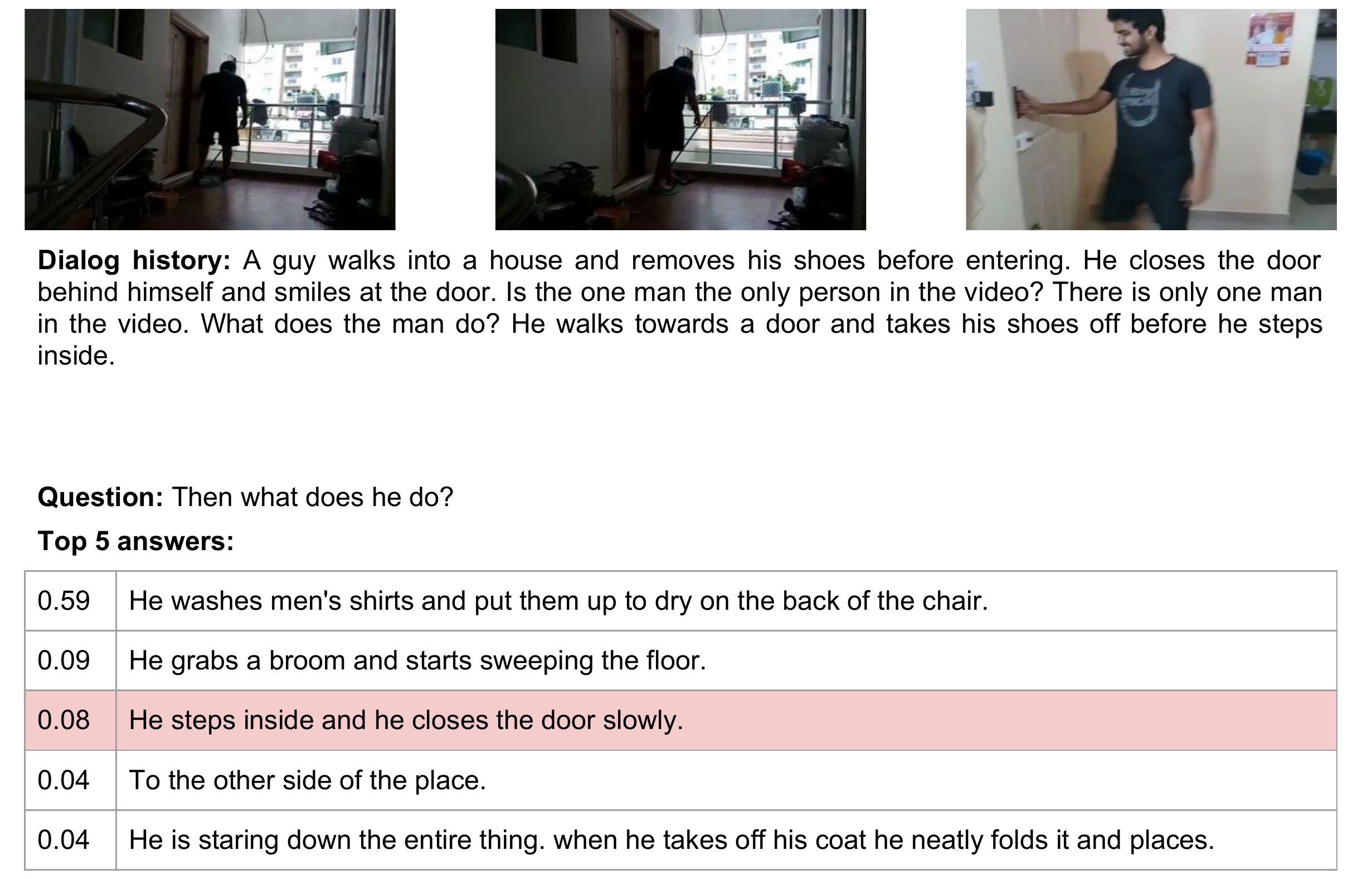}
	    \caption{ }
	    \label{fig:s_f}
	\end{subfigure}
	
    \vspace{6pt}
	\begin{subfigure}[b]{0.45\textwidth}
	    \centering
	    \includegraphics[width=\linewidth]{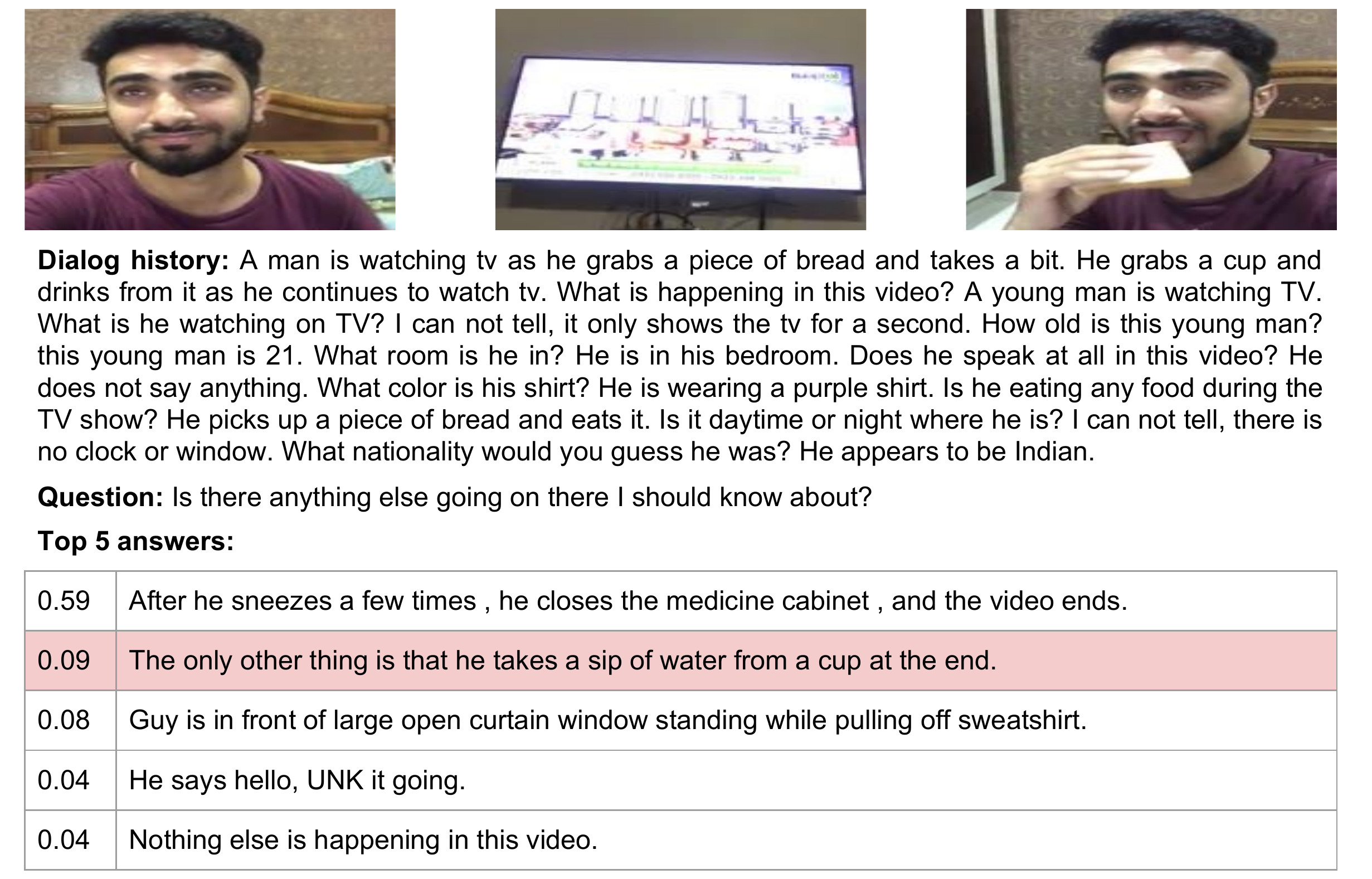}
	    \caption{ }
	    \label{fig:s_g}
	\end{subfigure}
		\hspace{10pt}
	\vrule
	\hspace{10pt}
	\begin{subfigure}[b]{0.45\textwidth}
	    \centering
	    \includegraphics[width=\linewidth]{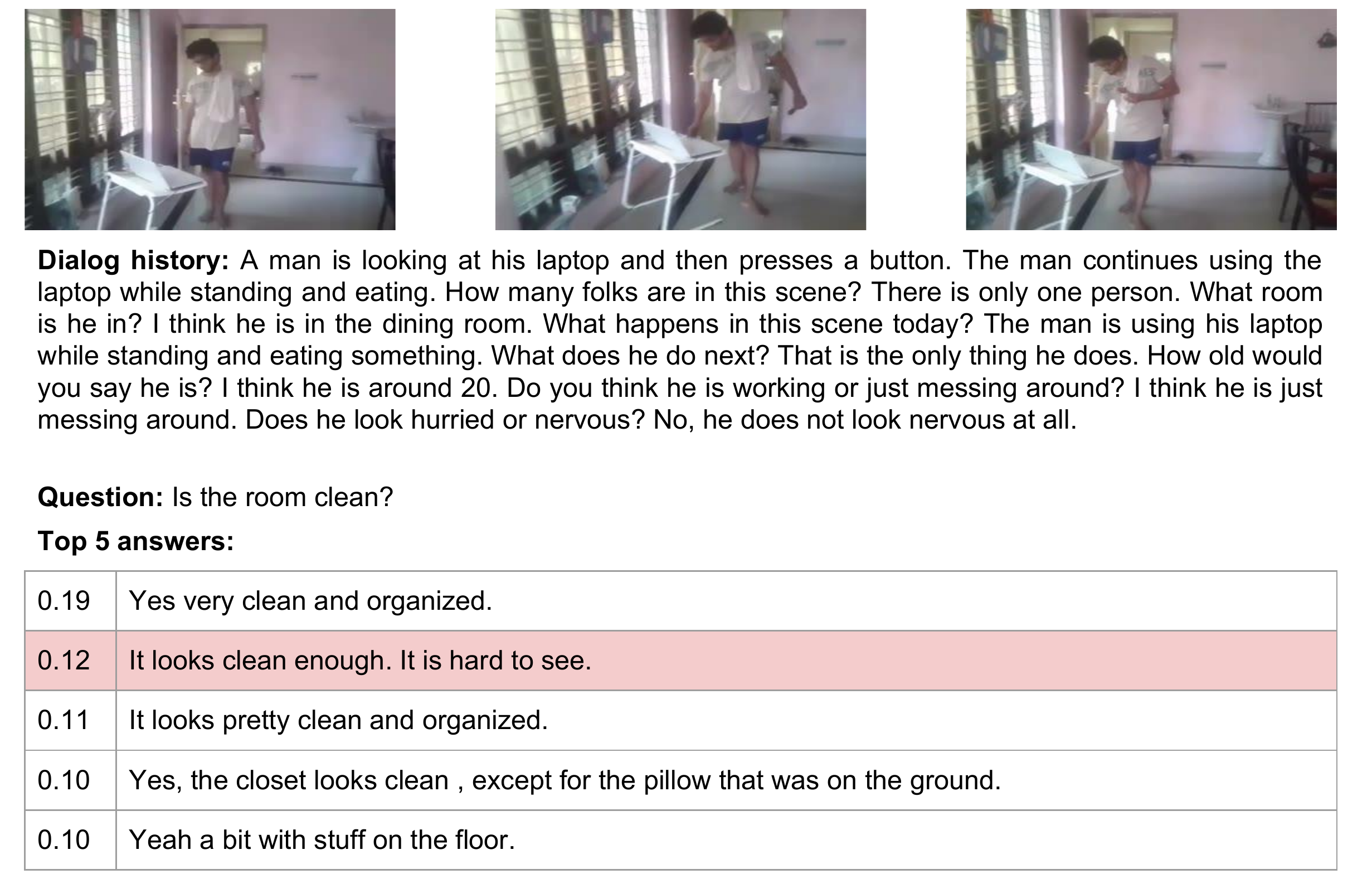}
	    \caption{ }
	    \label{fig:s_h}
	\end{subfigure}
	\caption{Examples using \texttt{Q+DH+V+A}. The left column of the tables in each figure represents the corresponding answer probability. The red highlights the ground truth answer.}
    \label{fig:qualitative_samples}
    \vspace{-8pt}
\end{figure*}

Figures \ref{fig:s_a} and \ref{fig:s_b} show examples of audio related questions. In figure \ref{fig:s_a}, although the model ranked the ground truth answer at third position, the two top ranked answers can also be valid answers to the given question "Dose he say any thing?". In \ref{fig:s_b}, three out of the top four ranked answers can be a valid answers as well. They all answered 'no' to the question. This highlights the deep understanding of the question and context.
Figures \ref{fig:s_c}, \ref{fig:s_d} and \ref{fig:s_e} are examples of visual-related questions. In figure \ref{fig:s_e}, the model must determine a person's age by leveraging visual cues from the video frames. An important type of question in video-based dialog is the temporal-based question. Examples of this type are shown in Figure \ref{fig:s_d} and \ref{fig:s_f}. Figures \ref{fig:s_g} and \ref{fig:s_h} show interesting and challenging questions about the general scene. In our dataset, there are no one-word answers such as "yes" or "no". The Answerers were asked to provide further details about their responses.

\subsection{Examples comparing setups \texttt{Q}, \texttt{Q+V}, \texttt{Q+A} and \texttt{Q+DH+V+A}}
\begin{figure*}[t]
\begin{center}
	\begin{subfigure}[b]{0.45\textwidth}
	    \centering
	    \includegraphics[width=\linewidth]{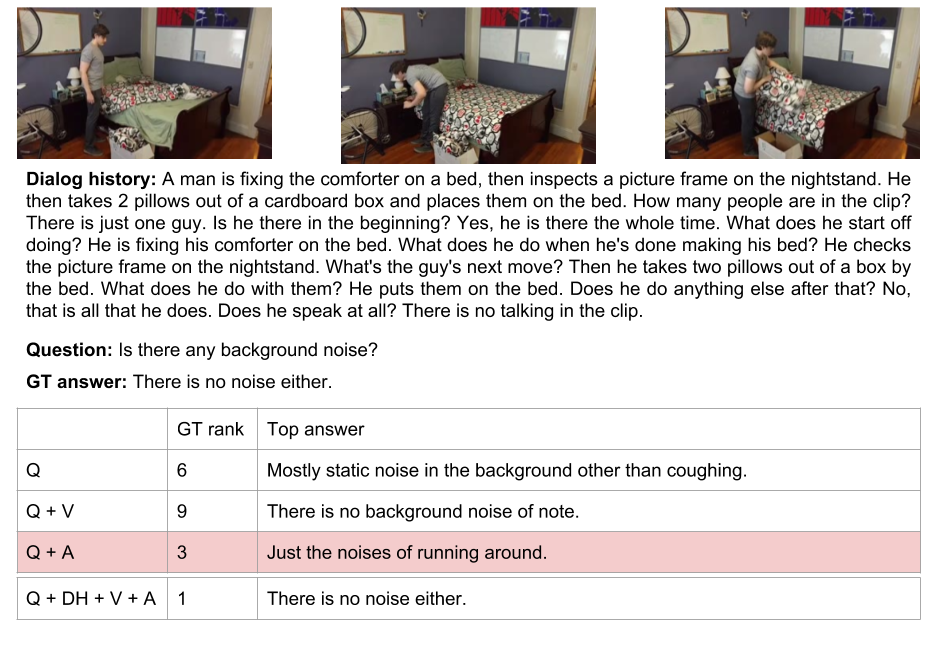}
	    \caption{ }
	    \label{fig:c_a}
	\end{subfigure}
	\hspace{10pt}
	\vrule
	\hspace{10pt}
	\begin{subfigure}[b]{0.45\textwidth}
	    \centering
	    \includegraphics[width=\linewidth]{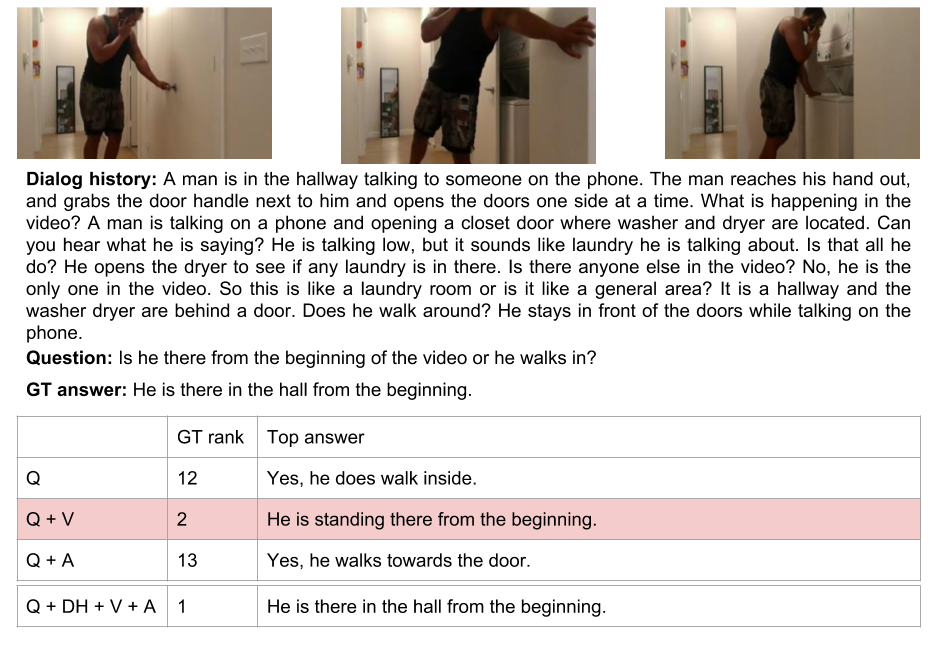}
	    \caption{ }
	    \label{fig:c_b}
	\end{subfigure}%
	
    \vspace{6pt}
	\begin{subfigure}[b]{0.45\textwidth}
	    \centering
	    \includegraphics[width=\linewidth]{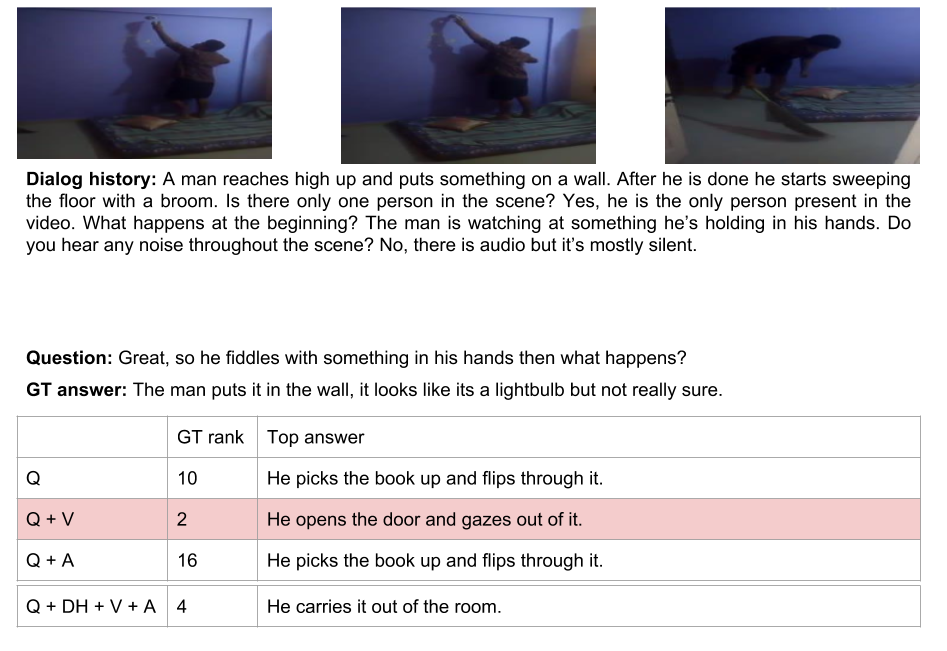}
	    \caption{ }
	    \label{fig:c_c}
	\end{subfigure}
\end{center}
    \vspace{-12pt}
	\caption{Comparison between models \texttt{Q}, \texttt{Q+V}, \texttt{Q+A} and \texttt{Q+DH+V+A}. The GT rank is the rank of the ground truth answer for the corresponding model. The top answer is the first ranked answer for the corresponding model. The red highlights the best model.}
    \label{fig:comp}
    \vspace{-8pt}
\end{figure*}

Figure \ref{fig:comp} shows examples comparing results between models \texttt{Q}, \texttt{Q+V}, \texttt{Q+A} and \texttt{Q+DH+V+A}. The GT rank is the rank of the ground truth answer for the corresponding model. The top answer is the first ranked answer for the corresponding model. The red highlights the best model. In figure \ref{fig:c_a} the question is audio related question and the \texttt{Q+A} model performs better. The question from the example in figure \ref{fig:c_b} is visual related question and the \texttt{Q+V} model performs best. Figure \ref{fig:c_c} presents a temporal related question best answered by the \texttt{Q+V} model. This highlights the value of each modality in the dataset.
\vspace{10pt}

\section{Summaries Interface.}

\label{sec:section_summaries_interface}

\begin{figure*}[t]
    \centering
    \vspace{10pt}
    \includegraphics[width=0.55\textwidth]{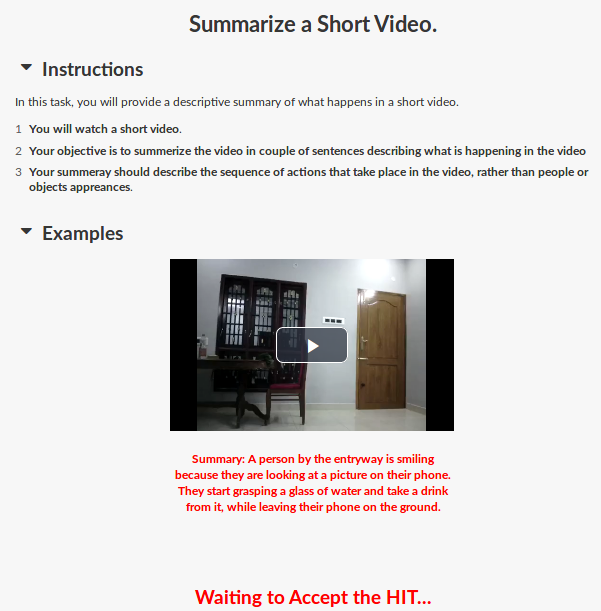}
    	\caption{Summaries data collection interface on AMT.}
 \vspace{-10pt}
\label{fig:summaryInterface}
\end{figure*}

The data collection process of AVSD included a downstream task, where  the Questioners had to write a summary of what they think happened in the video based on the conversation they had about it. To evaluate the quality of these conversations, we ran a separate study case on AMT. We asked 4 people to watch the video and write a summary describing all the events in the video. Figure \ref{fig:summaryInterface} shows the interface for this task. People where presented with example of the video and the script for that video. We then compared these summaries with the one written by the Questioner. 

\begin{figure*}[ht]
\begin{center}
	\begin{subfigure}[b]{0.45\textwidth}
	    \centering
	    \includegraphics[width=\linewidth]{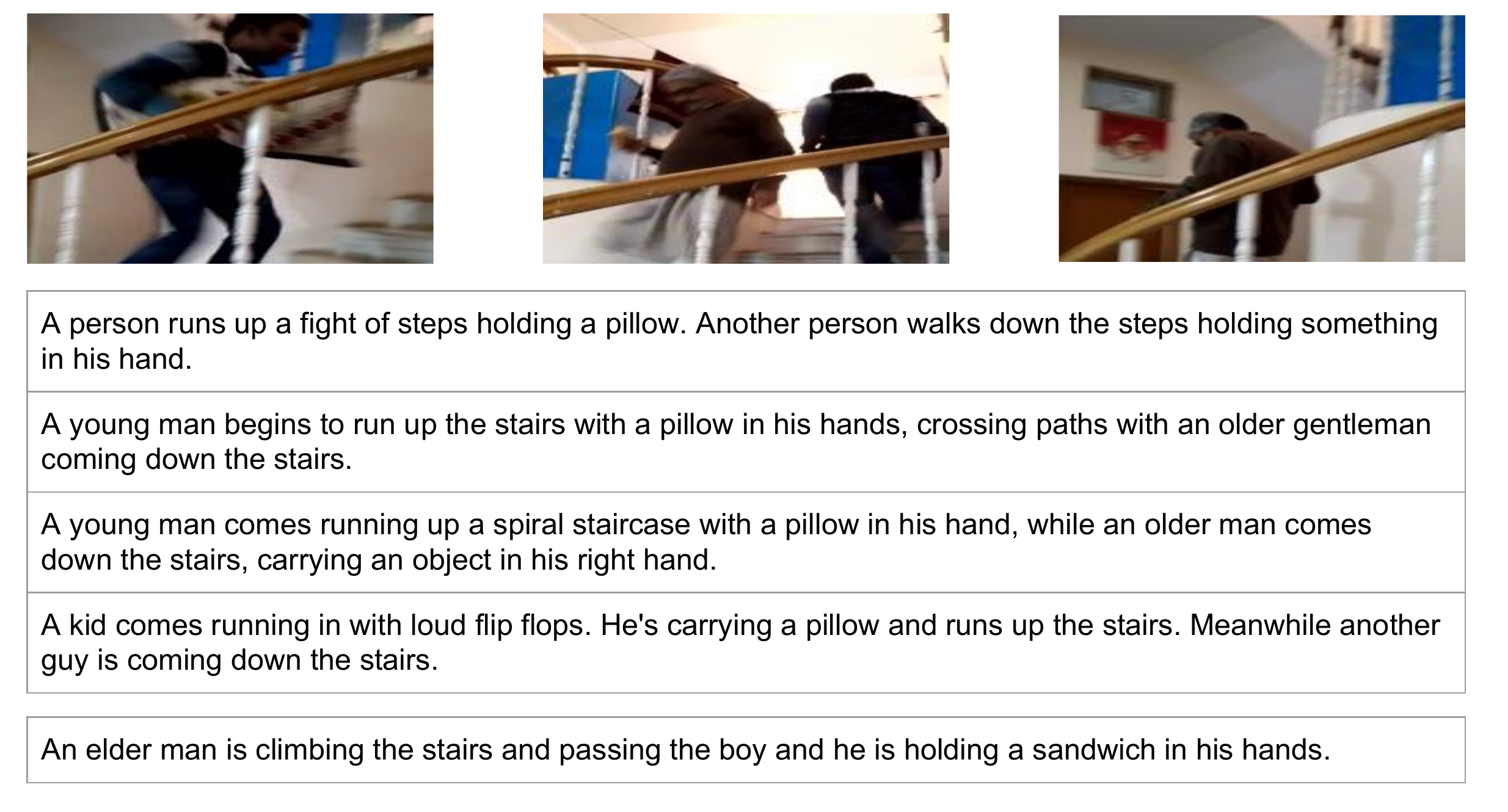}
	    \caption{ }
	    \label{fig:su_a}
	\end{subfigure}
	\hspace{10pt}
	\vrule
	\hspace{10pt}
	\begin{subfigure}[b]{0.45\textwidth}
	    \centering
	    \includegraphics[width=\linewidth]{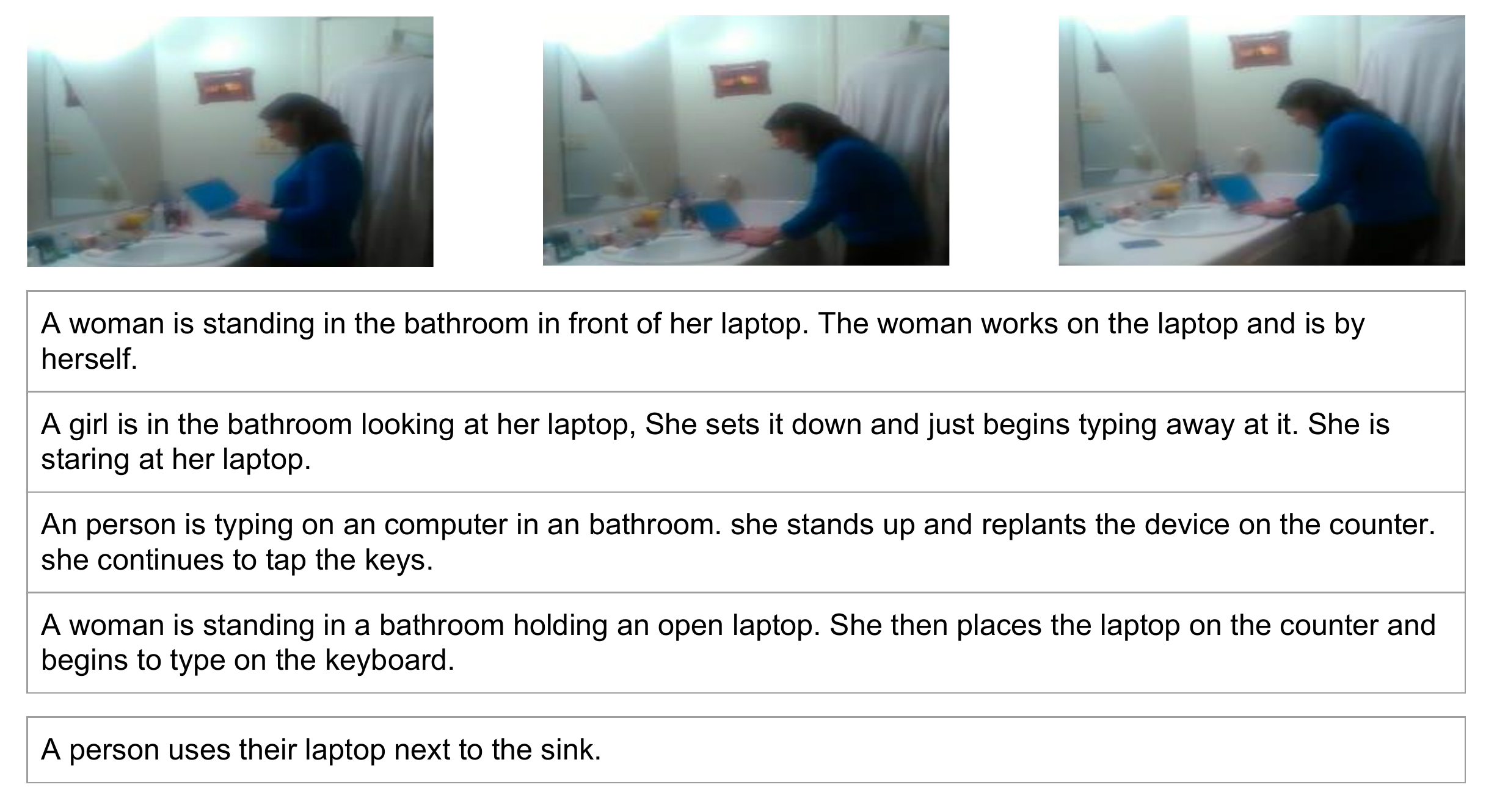}
	    \caption{ }
	    \label{fig:su_b}
	\end{subfigure}%
	
    \vspace{6pt}
	\begin{subfigure}[b]{0.45\textwidth}
	    \centering
	    \includegraphics[width=\linewidth]{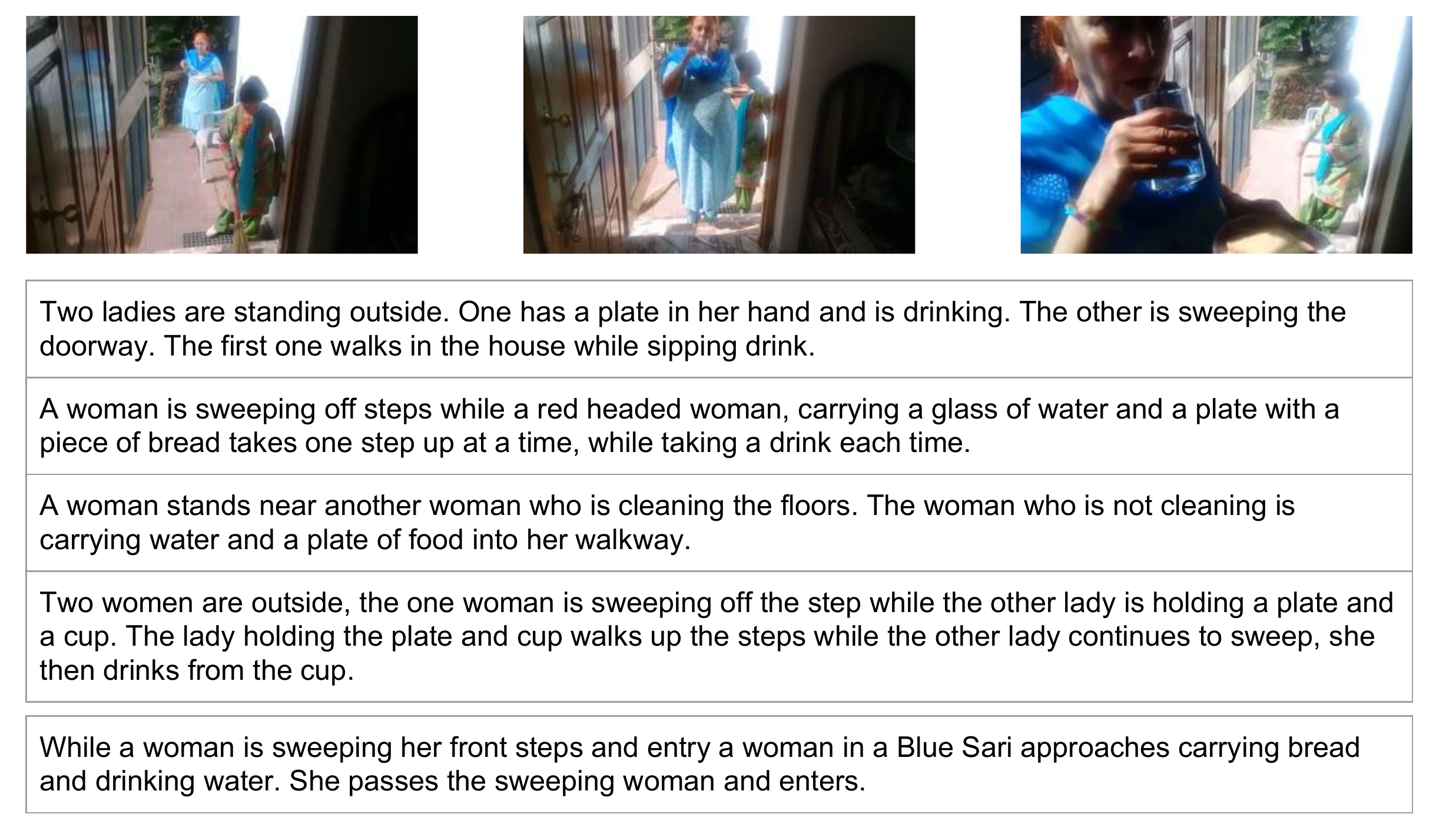}
	    \caption{ }
	    \label{fig:su_c}
	\end{subfigure}
\end{center}
    \vspace{-12pt}
	\caption{Comparison between different video summaries. The first 4 rows are summaries written by people after watching the entire video. Last row is summary written by the questioner who did not watch the video.}
    \label{fig:summaries}
    \vspace{-8pt}
\end{figure*}

Figure \ref{fig:summaries} shows some examples of the 4 summaries collected in this study (first four rows) and the summary written by the Questioner at the end of the dialog (last row). In these examples, we see that the summary written by the Questioner captures most of the events described in the 4 summaries.

\end{document}